\documentclass[10pt]{article} % For LaTeX2e
\usepackage[preprint]{tmlr}

\usepackage{amsmath,amsfonts,bm}

\def\eqref#1{equation~\ref{#1}}
\def\1{\bm{1}}

\DeclareMathAlphabet{\mathsfit}{\encodingdefault}{\sfdefault}{m}{sl}
\SetMathAlphabet{\mathsfit}{bold}{\encodingdefault}{\sfdefault}{bx}{n}

\usepackage{amsthm}

\newtheorem{lemma}{Lemma}
\newtheorem{theorem}{Theorem}

\newtheorem*{proposition*}{Proposition}

\newtheorem*{claim*}{Claim}
\theoremstyle{definition}

\newtheorem{definition}{Definition}
\newtheorem*{definition*}{Definition}

\usepackage{hyperref}
\usepackage{url}
\usepackage{nicefrac}       % compact symbols for 1/2, etc.
\usepackage{microtype}      % microtypography
\usepackage{graphicx}       % Required for inserting images
\usepackage[utf8]{inputenc}
\usepackage{amsfonts}
\usepackage{xcolor}         % colors
\usepackage{subcaption}
\usepackage{wrapfig}
\usepackage{bbm}
\usepackage{amssymb}% http://ctan.org/pkg/amssymb
\usepackage{bm} % bold math
\usepackage{pifont}% http://ctan.org/pkg/pifont
\usepackage{tikz}
\usepackage{colortbl}

\usepackage{booktabs}       % professional-quality tables
\usepackage{multirow}
\usepackage{outlines}

\usepackage{pgfplots}
\usepgfplotslibrary{groupplots}
\pgfplotsset{compat=newest}
\usetikzlibrary{pgfplots.fillbetween}% 
\usetikzlibrary{patterns, patterns.meta}
\pgfplotsset{compat=newest}
\usetikzlibrary{backgrounds,arrows.meta,hobby}
\usetikzlibrary{arrows, decorations.markings}
\usepgfplotslibrary{colorbrewer}

\usepackage{tikz-3dplot}
\colorlet{estColor}{green!40!black}
\colorlet{errColor}{red}
\colorlet{ubColor}{blue}
\colorlet{daubColor}{green!70!black}
\colorlet{legColor}{blue!80!black}
\colorlet{fouColor}{yellow!70!red}
\tikzset{>=latex} % for LaTeX arrow head
\tikzstyle{vector}=[-stealth,thick,line cap=round]
\usepackage{pgfplotstable}
\usetikzlibrary{decorations.pathreplacing,calligraphy}
\usetikzlibrary{fit, calc, positioning}

\title{ SaFARi: State-Space Models for Frame-Agnostic Representation}

\author{\name Hossein Babaei \email hb26@rice.edu \\
      \addr Department of Electrical and Computer Engineering\\
      Rice University
      \AND
      \name Mel White \email mel.white@rice.edu \\
      \addr Department of Electrical and Computer Engineering\\Rice University
      \AND
      \name Sina Alemohammad \email sa86@rice.edu \\ \addr Department of Electrical and Computer Engineering\\Rice University
      \AND
      \name Richard G. Baraniuk \email richb@rice.edu \\ \addr Department of Electrical and Computer Engineering\\Rice University
      }

\def\month{04}  % Insert correct month for camera-ready version
\def\year{2025} % Insert correct year for camera-ready version
\def\openreview{\url{https://openreview.net/forum?id=XXXX}} % Insert correct link to OpenReview for camera-ready version

\usepackage{amsthm}
\theoremstyle{plain}

\newtheorem*{apptheorem}{Theorem}
\newtheorem*{appprop}{Proposition}
\begin{document}

\maketitle

\begin{abstract}
State-Space Models (SSMs) have re-emerged as a powerful tool for online function approximation, and as the backbone of machine learning models for long-range dependent data.
However, to date, only a few polynomial bases have been explored for this purpose, and the state-of-the-art implementations were built upon the best of a few limited options.
In this paper, we present a generalized method for building an SSM with any frame or basis, rather than being restricted to polynomials.
This framework encompasses the approach known as HiPPO, but also permits an infinite diversity of other possible ``species'' within the SSM architecture.
We dub this approach SaFARi: SSMs for Frame-Agnostic Representation.
\end{abstract}

\section{Introduction}

Modeling sequential data is a cornerstone of modern machine learning, with applications spanning natural language processing, speech recognition, video analysis, and beyond \citep{eventcam2024,alemohammad2019wearing,S4ND_2022}. 
A fundamental challenge in these domains is the efficient representation of long-range dependence in time-series data, where the goal is to capture and preserve the essential features of the input signal necessary for downstream tasks over extended time horizons while maintaining computational tractability \citep{hochreiter1997long}. 

Machine learning approaches, such as recurrent neural networks (RNNs), have struggled with learning long-range dependencies due to challenges like vanishing gradients, exploding gradients, and limited memory horizons \citep{ElmanRNN,hochreiter1997long, schuster1997bidirectional,pascanu13}. 
During backpropagation through time, gradients in RNNs are repeatedly multiplied by the same weight matrix, causing them to either shrink exponentially (vanish) or grow exponentially (explode). 
Vanishing gradients prevent the network from updating weights effectively, while exploding gradients lead to unstable training. 
These issues hinder RNNs from maintaining and utilizing information over extended sequences. 
Although variants like LSTMs \citep{graves_lstm_2005} and GRUs \citep{gru} address some of these limitations, they often require task-specific parameterization and struggle to generalize across different sequence lengths or timescales.

State-space models (SSMs) have re-emerged as a powerful alternative for online representation of sequential data. 
By design, state-space models enable the online computation of compressive representations, maintaining a constant-size memory footprint regardless of sequence length.
The seminal work of \citeauthor{gu2020hippo} introduced High-Order Polynomial Projection Operators (HiPPO), which leverages orthogonal function bases to enable theoretically grounded, real-time updates of sequence representations. 
This framework, and its subsequent extensions such as S4 and Mamba, have demonstrated remarkable performance in tasks involving long-range dependencies, such as language modeling and signal processing \citep{gu2023mamba,gu2022s4,gu2022train,diag_effective_2024,diag_init_2022,smith2023s5,liquid_s4_2023}. 
By formulating sequence representation as an online function approximation problem, HiPPO provides a unified perspective on memory mechanisms, offering both theoretical guarantees and practical efficiency. 

However, despite its successes, the HiPPO framework has been limited to specific families of function bases, primarily orthogonal polynomials and Fourier bases.
While these bases are well-suited for certain applications, they are not universally optimal; 
different bases have varying levels of accuracy when representing different signal classes.
Fourier bases, for instance, are optimal for smooth, periodic signals due to their global frequency representation. 
Polynomial bases, such as orthogonal polynomials (e.g., Legendre or Chebyshev), are particularly effective for approximating smooth functions over compact intervals. 

The absence of a more flexible basis selection restricts the adaptability of the HiPPO framework.
In this work, we address this restriction by presenting a generalized method for constructing SSMs using any frame or basis of choice, which we term SaFARi (SSMs for Frame-Agnostic Representation). 
Our approach extends and generalizes the HiPPO framework using a numerical (as opposed to closed-form) method, which enables us to relax the requirements on the basis (such as orthogonality of the components).

Our key contributions are as follows:
\begin{itemize}
    \vspace{-0.1cm}
    \item \textbf{Generalized SSM construction:} We present SaFARi, a frame-agnostic method for deriving SSMs associated with any basis or frame, generalizing the HiPPO framework to a broader class of function representations.
    
    \item \textbf{Error Analysis:} We provide a comprehensive discussion of SaFARi's error sources and derive error bounds, offering theoretical insights into its performance and limitations.
\end{itemize}
\vspace{-0.3cm}
This paper is organized as follows. 
In Section \ref{sec:background}, we review the HiPPO framework and its limitations, motivating the need for a generalized approach. 
Section \ref{sec:math_prelim} provides the required mathematical preliminaries. 
Section \ref{sec:SAFARI} introduces our frame-agnostic method for SSM construction, and then  
Section \ref{sec:implementation} addresses the implementation considerations and strategies for SaFARi, including the approximation of its infinite-dimensional representation in finite dimensions, and provides a rigorous theoretical analysis of the associated errors. 
Finally, Section \ref{sec:conclusions} discusses the broader implications of our work and outlines directions for future research.

\section{Background}\label{sec:background}

Recent advances in machine learning, computer vision, and LLMs have exploited the ability to collect and contextualize more and more data over longer time frames. 
Handling such sequential data presents three main challenges: 1) generating a compact and low-dimensional representation of the sequence, 2) effectively preserving information within that representation, and 3) enabling real-time updates for streaming data. 

The classic linear method of obtaining the coefficients of a compressed representation of a signal is through a transform (e.g. Fourier) \citep{oppenheim1999discrete,abbate2012wavelets,box2015time,proakis2001digital,prandoni2008signal}.
However, a significant limitation of this method is its inefficiency in handling real-time updates. 
When new data arrives, the representation must be recalculated in its entirety, necessitating the storage of all prior data, which is sub-optimal in terms of both computation and storage requirements. 
This limits the horizon of the captured dependencies within the sequence.

Nonlinear models, such as recurrent neural networks (RNNs) and their variants, have been introduced more recently \citep{ElmanRNN,hochreiter1997long,gru,schuster1997bidirectional}.
Since these learned representations are task-specific, they are not easily utilized for other circumstances or applications.
Furthermore, RNNs struggle to capture long-range dependencies due to issues such as vanishing and exploding gradients.

\subsection{State-space models}

The state-space representation itself is not new; it was introduced by \citet{kalman} via the eponymous Kalman Filter.
For an input $u(t)$, output $y(t)$, and a state representation called $x(t)$, many systems and their properties can be described and controlled with the following system of linear equations:
\begin{align}
\begin{split}\label{eq:ssm}
    \dot{x}(t) &= Ax(t) + Bu(t) \\
    y(t) &= Cx(t) + Du(t).
\end{split}
\end{align}
In many classic applications, we iteratively update the matrices $A$, $B$, $C$, and $D$ to control or predict the output $y$ based on previous values of $u$.

\vspace{-0.2cm}
For online function approximation, however, we instead \textit{define} the matrices $A$ and $B$ such that they iteratively update a vector of coefficients $c$ over a particular basis.
For the moment, we can ignore $C$ and $D$  (or, equivalently, consider $C$ to be an identity matrix and $D=0$).
For stability, $A$ must have only negative eigenvalues, so we explicitly include a negative sign here. 
$A$ and $B$ may or may not be constant over time, so for completeness, we call these $A(t)$, $B(t)$.
Eq.~\ref{eq:ssm} is now
\vspace{-0.1cm}
\begin{equation}\label{eq:basic_ssm}
    \dot{c} = -A(t)c(t) + B(t)u(t).
\end{equation}
 \begin{figure}[t!]
     \centering
     \resizebox{0.6\textwidth}{!}{%}
     % \documentclass[10pt]{standalone}

% \usepackage{tikz}
% \usepackage{pgfplots, pgfplotstable}
% \usetikzlibrary{positioning, fit, shapes.geometric, bending, decorations.text, matrix}
% \usetikzlibrary{pgfplots.groupplots}
% \usepgfplotslibrary{colorbrewer}
% \pgfplotsset{compat=1.15}
% \usetikzlibrary{positioning,shapes,shadows,arrows}
% \usetikzlibrary{backgrounds}
% \usetikzlibrary{calc}

% \definecolor{cBlue}{HTML}{6495ED}
% \definecolor{cRed}{HTML}{CD5C5C}
% \definecolor{cPurple}{HTML}{9370DB}
% \definecolor{cGreen}{HTML}{2E8B57} %{097969}
% \definecolor{cOrange}{HTML}{ff7043}
% \definecolor{cYellow}{HTML}{FFFF8F}

% \definecolor{textColor}{HTML}{000000}
% \def\opacity{70}
% \def\xdist{0.75cm}
% \def\ydist{0.1cm}
% \def\rectheight{0.75cm}

% \begin{document}
\begin{tikzpicture}

\def\xdist{1cm};
\def\ydist{0.5cm};
\def\wline{0.3mm}
\definecolor{textColor}{HTML}{000000}
\def\opacity{70}
\def\rectheight{0.75cm}
\definecolor{cBlue}{HTML}{6495ED}
\definecolor{cRed}{HTML}{CD5C5C}
\definecolor{cPurple}{HTML}{9370DB}
\definecolor{cGreen}{HTML}{2E8B57} %{097969}
\definecolor{cOrange}{HTML}{ff7043}
\definecolor{cYellow}{HTML}{FFFF8F}

\node[rectangle, line width = \wline, text = textColor,
    minimum width = .5 cm, minimum height=\rectheight]
    (fk) {$u(t)$};

\node[circle, draw, line width = \wline, text = textColor,
    fill= white,
    font=\huge, 
    inner sep=0pt,
    minimum width = 0.5 cm, %minimum height=\rectheight,
    right = \xdist of fk.east]
    (times) {$\times$};

\node[circle, draw, line width = \wline, text = textColor,
    fill= white,
    font=\huge, 
    inner sep=0pt,
    minimum width = 0.5 cm, %minimum height=\rectheight,
    right = \xdist of times.east]
    (plus) {$+$};

\node[rectangle, draw, line width = \wline, text = textColor,
    fill = cBlue!\opacity, 
    minimum width = 2 cm, minimum height=\rectheight,
    right = \xdist of plus.east]
    (ode) {ODE Update};

\node[circle, draw, %cGreen!\opacity, 
    fill=black,
    inner sep=0pt,
    minimum width = 1.5mm, %minimum height=\rectheight,
    right = \xdist of ode.east]
    (ck) {};

\node[right=\xdist of ck.east] (out) {$c(t)$};

% \node[circle, draw, line width = \wline, text = textColor,
%     fill=white,
%     minimum width = 0.5 cm, minimum height=\rectheight,
%     right = 2*\xdist of ck.east]
%     (mult) {$\times$};

% \node[rectangle, line width = \wline, text = textColor,
%     minimum width = 1 cm, minimum height=\rectheight,
%     right = \xdist of mult.east]
%     (f_est) {$\tilde{f}$[0:k]};

% \node[rectangle, draw, line width = \wline, text = textColor,
%     fill = cOrange!\opacity,
%     minimum width = 1 cm, minimum height=\rectheight,
%     below = \ydist of mult.south]
%     (E) {E};

\node[circle, draw, line width = \wline, text = textColor,
    fill= white,
    font=\huge, 
    inner sep=0pt,
    minimum width = 0.5 cm, %minimum height=\rectheight,
    below = \ydist of plus.south]
    (times2) {$\times$};

\node[circle, draw, %cGreen!\opacity, 
    fill=black,
    inner sep=0pt,
    anchor=center,
    minimum width = 1.5mm, %minimum height=\rectheight,
    below = .77cm of ode.south]
    (ck2) {};

\node[rectangle, draw, line width = \wline, text = textColor,
    fill = cRed!\opacity, 
    minimum width = 1 cm, minimum height=\rectheight,
    below = \ydist of times2.south]
    (A) {$-A$};

\node[rectangle, draw, line width = \wline, text = textColor,
    fill = cRed!\opacity, 
    anchor=north,
    minimum width = 1 cm, minimum height=\rectheight,
    below = 1.75cm of times.south]
    (B) {$B$};

\draw[line width=\wline,->] (fk.east) -- (times.west);
\draw[line width=\wline,->] (times.east) -- (plus.west);
\draw[line width=\wline,->] (B.north) -- (times.south);
%label={[<options>]<position>:<text>}
\draw[line width=\wline,->] (plus.east) -- (ode.west) node [midway, label={[yshift=-1mm]above:{$\dot{c}(t)$}}] {};   
\draw[line width=\wline,->] (ode.east) -- (out.west);   
\draw[line width=\wline,->] (ck.south) |- (times2.east);
\draw[line width=\wline,->] (times2.north) -| (plus.south);
\draw[line width=\wline,->] (A.north) -- (times2.south);
\draw[line width=\wline,->] (ck2.north) -- (ode.south);
%\draw[line width=\wline,->] (ck.east) -- (out.west);  
% \draw[line width=\wline,->] (E.north) -- (mult.south);
% \draw[line width=\wline,->] (mult.east) -- (f_est.west);  

\begin{scope}[on background layer]
\node[draw,inner xsep=10pt,inner ysep=10pt, 
    fill=gray!30,
    fit={(B)(ode)(ck)(A)(plus)},label={above:{SSM}}] (SSM) {};
%\node[draw,inner xsep=10pt,inner ysep=10pt, 
%    fill=gray!30,
%    fit={(E)(mult)},label={above:{Reconstruction}}](recon){};
\end{scope}
\end{tikzpicture}

% \end{document}
     }
     \vspace{-0.2cm}
     \caption{An SSM block-diagram, with the necessary ODE update step included. }
     \label{fig:bkd}
\end{figure}
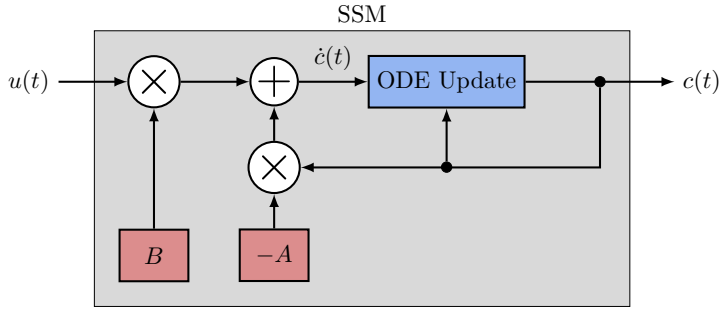
The challenge in the problem of the approximation of online functions is to derive appropriate matrices $A$ and $B$, which was answered by HiPPO \citep{gu2020hippo}.

\subsection{HiPPO: High-order Polynomial Projection Operators}\label{sec:hippo}
The HiPPO framework \citep{gu2020hippo} enables online function approximation using pre-determined SSM parameters derived from a basis of orthogonal polynomials $\mathcal{G}$. 
It optimizes $\| u_{T} - g^{(T)}(t) \|_{\mu}$ for $g^{(T)}(t) \in \mathcal{G} $, to find a set of coefficients for the orthogonal polynomials at every time $T$, which yields the approximation of the input function over $[0,T]$.

\begin{figure}[b!]
   \centering
        \resizebox{\textwidth}{!}{%}
        % \documentclass[tikz]{standalone}
% \usepackage{tikz,pgfplots}
% \usetikzlibrary{backgrounds,arrows.meta,hobby}
% \usepgfplotslibrary{fillbetween}
% \usepgfplotslibrary{groupplots}
% \pgfplotsset{compat=1.18}

% \begin{document}

% \colorlet{shadeColor}{red}

% \def\cvals{(0,0) (0.1,0.15)  (0.2,0.3)  (0.3,0.22) (0.4,0.5) (0.5,0.58)  (0.6,0.43) (0.7,0.4) (0.8,0.38) (0.9,0.3)  (1,0.24)}

\begin{tikzpicture}[declare function={ f(\fcoords)= coordinates {\fcoords}; }]
    \begin{groupplot}[group style=
        {group size=3 by 2,     
        xlabels at=edge bottom,
        ylabels at=edge left,},
    height=3.5cm,width=7cm,
    xtick={0,0.2,0.6,1},
    xticklabels={$t_0$,$t_1$,$t_2$,$t_T$},
    ymin=0, ymax=1,
    xmin=0, xmax=1,
    ytick=2,
    ]
    \colorlet{shadeColor}{red}

    \def\cvals{(0,0) (0.1,0.15)  (0.2,0.3)  (0.3,0.22) (0.4,0.5) (0.5,0.58)  (0.6,0.43) (0.7,0.4) (0.8,0.38) (0.9,0.3)  (1,0.24)}

    \nextgroupplot[ylabel=Scaled ($\mu_{\textit{sc}}$)]
    \addplot [opacity=0, fill=shadeColor, fill opacity=0.5, ylabel={Scaling}] (0, 0) -- (0, 1) -- (0.2, 1) -- (0.2, 0) -- (0,0); 
    ];
    \addplot [name path=f, blue, smooth] coordinates {\cvals};
    \nextgroupplot
    \addplot [opacity=0, fill=shadeColor, fill opacity=0.3] (0, 0) -- (0, 1) -- (0.6, 1) -- (0.6, 0) -- (0,0); 
    ];
    \addplot [blue, smooth] coordinates {\cvals};
    \nextgroupplot
    \addplot [opacity=0, fill=shadeColor, fill opacity=0.1] (0, 0) -- (0, 1) -- (1, 1) -- (1, 0) -- (0,0); 
    ];
    \addplot [blue, smooth] coordinates {\cvals};
    \nextgroupplot[ylabel=Translated ($\mu_{\textit{tr}}$)]
    \addplot [opacity=0, fill=shadeColor, fill opacity=0.5] (0, 0) -- (0, 1) -- (0.2, 1) -- (0.2, 0) -- (0,0); 
    ];
    \addplot [blue, smooth] coordinates {\cvals};
    \nextgroupplot
    \addplot [opacity=0, fill=shadeColor, fill opacity=0.5] (0.4, 0) -- (0.4, 1) -- (0.6, 1) -- (0.6, 0) -- (0.4,0);
    ];
    \addplot [blue, smooth] coordinates {\cvals};
    \nextgroupplot
    \addplot[opacity=0, fill=shadeColor, fill opacity=0.5] (0.8, 0) -- (0.8, 1) -- (1, 1) -- (1, 0) -- (0.8,0);
    ];
    \addplot [blue, smooth] coordinates {\cvals};
    % \nextgroupplot[ylabel=Exp. Decay]
    % \shade[left color=shadeColor!10, right color=shadeColor,opacity=0.5] (0,0) rectangle (0.2,1);
    % ];
    % \addplot [blue, smooth] coordinates {\cvals};
    % \nextgroupplot
    % \shade[left color=shadeColor!10, right color=shadeColor,opacity=0.5] (0,0) rectangle (0.6,1);
    % ];
    % \addplot [blue, smooth] coordinates {\cvals};
    % \nextgroupplot
    % \shade[left color=shadeColor!10, right color=shadeColor,opacity=0.5] (0,0) rectangle (1,1);
    % ];
    % \addplot [blue, smooth] coordinates {\cvals};
    \end{groupplot}

\end{tikzpicture}

% \end{document}
    }
    \vspace{-0.2cm}
    \caption{Two different uniform measures (red) applied to a signal (blue). The red shaded area demonstrates how the measure changes as time evolves and more samples of the input are observed. }
    \label{fig:measures}
\end{figure}
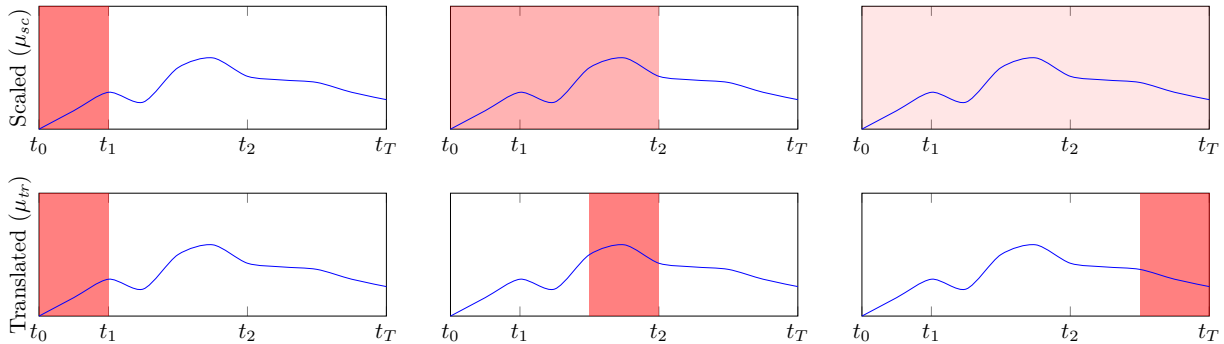

In addition to choosing the set of polynomials, one must also select a measure: the combination of a weighting function and a windowing function.
The window indicates which samples are included, and the weighting assigns a weight to each sample.
HiPPO \citep{gu2020hippo} considered several possible measures, two of which (illustrated in Fig.~\ref{fig:measures}) are:
\begin{equation}
    \mu_{\textit{tr}}(t)= \frac{1}{\theta}\mathbbm{1}_{t\in[T-\theta,T]} ,\quad \mu_{\textit{sc}}(t)= \frac{1}{T}\mathbbm{1}_{t\in[0,T]}. \vspace{-0.2cm}  \label{Measures}
\end{equation}
The uniform translated measure, $\mu_{\textit{tr}}(t)$, gives equal weight to all samples in the most recent window with a constant length $\theta$, and zero weight to previous samples.
The uniform scaled measure, $\mu_{\textit{sc}}(t)$, gives equal weight to all the times observed since $t=0$.  
This can be interpreted as squeezing or stretching the basis or frame to match the current length of the signal at any given time. Thus, the representation produced by this measure becomes less informative about the finer details of the signal as more history is observed, since the stretching of the basis will gradually increase the lowest observable frequency. 
Gu et al.\ also considered another weighting function, $\mu_{\textit{ed}}(t)$, which is a translated exponential decay measure that gives exponentially less weight to earlier parts of the signal history.
This work considers only uniformly weighted measures.  
We leave alternative weighting schemes to future work.

For the space of orthogonal polynomials, Hippo ODE update \citep{gu2020hippo} (see Fig.~\ref{fig:bkd}) follows a first-order linear differential equation: 
\vspace{-0.2cm}
\begin{equation}
   \frac{d}{dT} \vec{c}(T)= -A_{(T)} \vec{c}(T) +  B_{(T)} u(T) 
   \label{Time_Varying_SSM}
   \vspace{-0.2cm}
\end{equation}

where $u(T)$ is the value of the input signal at the current time $T$, and $\vec{c}(T)$ is the vector containing the representation of the $t \in [0,T]$ part of the input signal. $A_{(T)}$ and $B_{(T)}$ are also a time-varying matrix and vector that can be derived for the particular choice of the measure and orthogonal polynomial.

\textbf{Solving the differential equation to update the SSM}:
The differential equation (Eq.~\ref{Time_Varying_SSM}) can be solved incrementally, and there are several methods to do so.
We mention the \emph{generalized bilinear transform (GBT)} \citep{Zhang_GBT} and point the reader to \citet{butcher1987numerical} for other methods. 
The update rule given by the GBT method is
\begin{equation}
    \vspace{-0.1cm}
    c(t+\Delta t)=(  I  + \delta t \alpha A_{(t+ \delta t)} )^{-1} \, (I - \delta t (1-\alpha) A_{(t)} ) c(t) \, +\, (  I  + \delta t \alpha A_{(t+ \delta t)} )^{-1} \, \delta t B(t) u(t),
    \label{Update_GBT}
    \vspace{-0.1cm}
\end{equation}
where $0 \leq \alpha \leq 1$ is a chosen constant. 
(Note that for $\alpha=0$, the update rule becomes the same as the forward Euler method.)  
In this paper, we rely on findings from \citet{gu2020hippo} that the GBT produces the best numerical error in solving first-order SSMs.

\textbf{Diagonalizing the transition matrix \texorpdfstring{$\boldsymbol{A}$}{}}: 
The incremental update given by the GBT requires a matrix inversion and matrix products at each increment. 
Having a diagonal $A$ significantly reduces the computational cost of these operations. 
If the measure used for the SSM is such that the eigenvectors of $A(t)$ remain constant (for example, if $A(t)$ is a constant matrix multiplied by an arbitrary function of time), then it is possible to find a change of basis that makes the SSM matrix diagonal. 
To do this, one finds the eigenvalue decomposition of the matrix $A(t)= V \Lambda(t) V^{-1}$ and re-write the SSM as 
\begin{equation}
    \vspace{-0.2cm}
    \frac{\partial}{\partial t} \widetilde{c} = -  \Lambda(t) \widetilde{c} + \widetilde{B} u(t), 
    \label{Diagonal_SSM}
    \vspace{-0.2cm}
\end{equation}
where $\widetilde{c}=V^{-1}c$ and $\widetilde{B}=V^{-1} B(t)$. 
This means that one can solve Eq.~\ref{Diagonal_SSM} instead of Eq.~\ref{Time_Varying_SSM}, then find the representation $c$ from $\widetilde{c}$ with a single matrix multiplication. 

\subsection{Limitations of HiPPO}

The original HiPPO formulation and a subsequent follow-up \citep{gu2022train} included a handful of orthogonal polynomial bases with specific measures.
Measures were chosen heuristically for each basis, so there is no method in the literature for arbitrary choice of basis and measure (see Table~\ref{tab:implementations}).
Moreover, it was found that many of the bases explored in the early work actually perform quite poorly for function approximation tasks.  
Only one, the scaled-Legendre (LegS), empirically performed well, but it introduced additional challenges as its $A$ matrix is not stably diagonalizable \citep{diag_init_2022}. 

\begin{table}[t!]
\begin{center}

\begin{tabular}{c c c c} \toprule
& Measure      &  Scaled        & Translated            \\ \midrule
\multirow{7}{*}{\rotatebox[origin=c]{90}{Basis or Frame}} &
 \multirow{2}{*}{Legendre}   &  \citet{gu2020hippo},\S\ref{sec:SAFARI}  &  \citet{gu2020hippo},\S\ref{sec:SAFARI} \\ 
 &    &  *\citet{gu2022train} &  \\ \cline{3-4}
& Fourier    &   \citet{gu2020hippo}, \S\ref{sec:SAFARI}&   \citet{gu2020hippo},\S\ref{sec:SAFARI} \\ \cline{3-4}
& Laguerre   &  \S\ref{sec:SAFARI}  & **\citet{gu2020hippo},\S\ref{sec:SAFARI} \\ \cline{3-4}
& Chebychev  &  \S\ref{sec:SAFARI}  &  **\citet{gu2020hippo}\S\ref{sec:SAFARI}  \\  \cline{3-4}
& Arbitrary  &     \S\ref{sec:SAFARI}              &   \S\ref{sec:SAFARI}  \vspace{0.1cm}  \\  \bottomrule 
\end{tabular}
\vspace{-0.3cm}
\caption{An overview of the combinations of frames (or bases) and measures covered in the literature to date. 
SaFARi fills in missing combinations in this table for the scaled and translated measures in Section \ref{sec:SAFARI}.
Note that the S4 work and beyond (denoted by {*}), which is placed in this table under Uniform/Scaled, technically applies an exponential measure in the ODE; however, the $A$ matrix of the SSM is generated based on a uniformly weighted scaled Legendre polynomial basis. Some implementations in the translated case (denoted by {**}) also apply an exponential decay measure.  We consider only the case of uniform weighting for each measure for this foundational work, but note that there are infinite alternate weighting schemes for future work to explore. }
\label{tab:implementations}
\end{center}
\end{table}

The majority of the follow-up work since HiPPO has abandoned the task of function approximation by SSMs alone. 
Instead, most research has employed a diagonal approximation of the best-performing extant version (LegS) as an \textit{initialization} for machine learning architectures such as S4 \citep{gu2022s4, smith2023s5} and Mamba \citep{gu2023mamba}.  
Still, the HiPPO framework still holds untapped potential for online function approximation and better initializations for learning tasks.

\section{Mathematical Preliminaries}\label{sec:math_prelim}
\vspace{-0.2cm}
Prior to introducing SaFARi, we first cover some required theoretical background on the use of frames for function representation and approximation.
This section will cover distinctions between approximations performed on the full signal all at once, and approximation performed in a sequential (online) fashion.

\subsection{Function representation using frames}

Given a function $u(t)$ of a time domain $t\in \mathcal{D}_T$, we aim to represent $u(t)$ using a collection of functions over $ \mathcal{D}_T $. 
This representation is performed with respect to a measure that determines the relative importance of the function's value at different times.
We formulate the task of function representation using frames via the definitions below.

\textbf{Frame:} The sequence $\Phi=\{ \phi_n\}_{n \in \Gamma}$ is a frame in the Hilbert space $\mathcal{H}$ if there exist two constants $A>0$ and $B>0$ such that for any $f \in \mathcal{H}$:
\vspace{-0.2cm}
\begin{equation}
    A_{\rm{frame}} \| f \|^2 \leq \sum_{n \in \Gamma} | \langle  f, \phi_n \rangle  |^{2}  \leq B_{\rm{frame}} \| f \|^2 
    \label{Frame_condition}
\end{equation}
where $  \langle \cdot ,\cdot \rangle$ is the inner product of the Hilbert space $\mathcal{H}$, and $\Gamma$ denotes the indices of the frame elements in $\Phi$. 
If  $A_{\rm{frame}}=B_{\rm{frame}}$, then the frame is said to be tight \citep{mallat_tour, christensen2003introduction, grochenig2001foundations}.

\textbf{Frame operator:} If the sequence $\Phi=\{ \phi_n\}_{n \in \Gamma}$ is a frame, then its associated operator $ \mathbb{U}_\Phi$ is defined as:
\begin{equation}
    \mathbb{U}_\Phi f=\vec{x},\qquad  x_{n}= \langle  f, \phi_n  \rangle , \quad \forall n \in \Gamma.
    \label{Frame_operator}
\end{equation}
It can be shown that the frame condition (Eq.~\ref{Frame_condition}) is necessary and sufficient for the frame operator (Eq.~\ref{Frame_operator}) to be invertible on its image with a bounded inverse. This makes the frame a complete and stable (though potentially redundant) framework for signal representation. 

\textbf{Dual frame:}  The set $\widetilde{\Phi}=\{ \widetilde{\phi}_n\}_{n \in \Gamma}$ is the dual of the frame $\Phi$, $\widetilde{\Phi}= \text{Dual}\{ \Phi \} $ if:
\vspace{-0.2cm}
\begin{equation}
      \widetilde{\phi}_n= (\mathbb{U}_\Phi^* \mathbb{U}_\Phi )^{-1} \phi_n
    \label{Frame_dual}
\end{equation}
where $\mathbb{U}_\Phi^* $ is the adjoint of the frame operator:  $ \langle \mathbb{U}_\Phi f,x\rangle=\langle f,\mathbb{U}_\Phi^* x\rangle  $.

It can be shown that the composition of the dual frame and the frame is the identity operator in the Hilbert space $\mathcal{H}$:  $\mathbb{U}^*_{\widetilde{\Phi}} \mathbb{U}_\Phi f = f$ \citep{mallat_tour, christensen2003introduction, grochenig2001foundations}. 
Thus, we can think of the dual frame as the operator that transforms the signal from frame representation back into the signal space. 

\textbf{Function approximation using frames:} The compressive power of frame-based function approximation lies in its ability to efficiently represent functions using a relatively small number of frame elements. 
Different classes of functions exhibit varying effectiveness in capturing the essential features of different signal classes. 
This efficiency is closely tied to the decay rate of frame coefficients, which can differ significantly between frames for a given input function class. 
As a result, selecting an appropriate frame is critical for optimal approximations while minimizing computational resources and storage space.

In the task of online function representation, we aim to represent a function $u_{T} := \{ u(t) , t \in [0,T]\}$ for any given time $T$, using a frame $\Phi$  that has the dual $\widetilde{\Phi}$ and the domain $t \in \mathcal{D}$. 
To do so, we need an invertible warping function so that the composition of the frame and the warping function $ \Phi^{(T)}$ includes the domain $[0,T]$. 
% Without loss of generality, we assume that the frame has the domain $\mathcal{D}_\Phi=[0,1]$ (If this is not the case, then one can easily find another warping function to warp $\mathcal{D}_\Phi \rightarrow [0,1]$ and apply it beforehand), and use the warping $\phi^{(T)}(t)= \phi\left(\frac{t}{T}\right)$. 
Without loss of generality, we assume that the frame has the domain $\mathcal{D}_\Phi=[0,1]$, and use the warping $\phi^{(T)}(t)= \phi\left(\frac{t}{T}\right)$. 
(If this is not the case, then one can easily find another warping function to warp $\mathcal{D}_\Phi \rightarrow [0,1]$ and apply it beforehand.)
To calculate the representation, the frame operator of $\Phi_T$ acts on $u_T$:
\begin{equation}
    \textit{Projection:} \quad x= \Phi^{(T)} u_T ,\quad  x_n= \langle u_T , \phi^{(T)}_n \rangle= \int_{t=0}^T u(t) \overline{  \phi^{(T)}_n(t) } \mu(t) dt,  
    \label{Analysis}
\end{equation}
where the overline in $\overline{\phi}(t)$ is the complex conjugate of $\phi(t)$. 
Then, the dual frame operator transforms the discrete representation back to the function $u_T$:
\begin{equation}
    \textit{Reconstruction:} \quad u_T= \widetilde{\Phi}^{(T)} \Phi^{(T)} u_T ,\quad  u_T(t)= \sum_{n\in \Gamma}  \langle u_T , \phi^{(T)}_n \rangle \widetilde{\phi}^{(T)}_n (t) \, .
    \label{Synthesis}
    \vspace{-0.2cm}
\end{equation}

\textbf{Measures:} Throughout this paper, we work with the Hilbert space consisting of all functions over the domain $\mathcal{D}$ with the inner product defined as:
\begin{equation}
    \langle u , v  \rangle = \int_{t\in \mathcal{D}} u(t) \overline{ v(t)}  \mu(t) dt.
    \label{eq:Inner_product}
\end{equation}
We will consider both scaling and translating windows with uniformly weighted measures as described in Eq.~\ref{Measures} and Fig.~\ref{fig:measures}, and derive the SSMs for each.
We do not implement weighting schemes other than uniform in this work explicitly, but any other weighting function can be implemented with our method by using Eq.~\ref{eq:Inner_product} and the appropriate SSM derivation for a scaled or translated window.  

\section{SaFARi}\label{sec:SAFARI}

Our primary contribution in this paper is a generalized and adaptable solution to the problem of online function representation using frames that we dub SaFARi. 
We show that the update rule for online approximation using a frame can be implemented as a 
first-order linear differential equation following a state-space model.
We provide our frame-agnostic solutions for the uniform translated and scaled measures. 
%\vspace{-0.2cm}
\subsection{Formulation}
%\vspace{-0.2cm}
We begin by presenting the formulation for the online function representation using a frame and determine the dynamics of its evolution. 
For the given frame $\Phi$ (without any loss of the generality, we assume has the domain $\mathcal D=[0,1]$), and an input function $u(T)$, the objective is to find the representation of the function using the frame $\Phi$ similar to Eq.~\ref{Analysis}:
\vspace{-0.2cm}
\begin{equation}
    \textit{Projection:} \quad  c_n(T) = (\mathbb{U}_{\Phi} u )_n= \int_{t=t_0}^T u(t) \overline{  \phi^{(T)}_n(t) } \mu(t) dt \, .
    \label{Analysis_WaLRUS}
\end{equation}
The representation vector $\vec{c}(T)$ is a vector with its $n^{\rm{th}}$ component defined as above.  We next find the derivative of the representation with respect to the current time $T$, and show that it follows a particular SSM for a scaled or translated measure.

\subsection{Uniform scaled measure \texorpdfstring{$\boldsymbol{\mu_{sc}}$}{}}

When we use the scaled measure ($\mu_{sc}(t)$ in Eq.~\ref{Measures}), the representation generated by applying the frame operator to the observed history of the input $u_T(t)$ is
\begin{equation}
    c_n(T) = \int_{t=0}^T u(t) \overline{  \phi_n } \left(\frac{t}{T}\right) \frac{1}{T} dt \, .
    \label{Scaling_representation}
\end{equation}

\begin{definition}
    For a given frame $\Phi$ consisting of functions on $\mathcal{D}$, we define the auxiliary derivative of the frame $\Upsilon_\Phi=\{ \upsilon_n  \}_{n\in{\Gamma}}$ as a collection having $\upsilon_n = t \frac{\partial}{\partial t} \phi_n(t) $. 
\end{definition}

The auxiliary derivative of the frame is the result of the operator $t \frac{\partial}{\partial t} $ acting on each individual frame element.  Note that the auxiliary derivative of the frame is not necessarily a frame itself and the frame condition in Eq.~\ref{Frame_condition} does not necessarily hold for $\Upsilon_\Phi$.

\begin{theorem}
\label{Theorem_Scaling_SSM}
For the representation defined in Eq.~\ref{Scaling_representation}, the partial derivative of $\vec{c}$ with respect to $T$ is
\begin{equation}
    \frac{\partial}{\partial T} \vec{c} (T) = - \frac{1}{T} A \vec{c} (T)+ \frac{1}{T} B u(T), \quad A= I+ \mathbb{U}_{\Upsilon}  \mathbb{U}^{*}_{\widetilde{\Phi}} 
    \label{Scaling_SSM}
\end{equation}
%where $A$ is the linear operator
%\begin{equation}
    %\overline{A}= I+ \mathbb{U}_{\Upsilon}  \mathbb{U}^{*}_{\widetilde{\Phi}} 
%\end{equation}
where $B$ is the complex conjugate of a vector containing members of the main frame evaluated at $T=1$ so that $B= \{ \overline{\phi}_n (T=1) \}_{n \in \Gamma}$. 
The $A$ operator can also be described as a matrix 
\begin{equation}
    A_{i,j}=\delta_{i,j} + \int_0^1 \left[ t \frac{\partial}{\partial t}\overline{\phi}_{i}(t) \right]_{t=t'} \widetilde{\phi}_{j} (t') dt' .
    \label{scaling_matrix}
\end{equation}
\end{theorem}
Proof is provided in the Appendix~\ref{Proof_scaling}. We will refer to the SSM with a scaling measure as scaled-SaFARi.  

\subsection{Uniform translated measure \texorpdfstring{$\boldsymbol{\mu_{tr}}$}{}}
When the translated  measure is used ($\mu_{tr}(t)$ in Eq.~\ref{Measures}), the representation resulting from applying the frame operator to the window of recently observed input $u_T(t)$ is
\begin{equation}
    c_n(T) = \int_{t=T-\theta}^T u(t) \overline{  \phi_n } \left(\frac{t-(T-\theta)}{\theta}\right) \frac{1}{\theta} dt \, .
    \label{Translated_representation}
\end{equation}
\begin{definition}
    For a given frame $\Phi=\{ \phi_n \}_{n \in \Gamma}$ consisting of functions of time, we define the time derivative of the frame $\dot{\Phi}=\{ \dot{\phi}_n = \frac{\partial}{\partial t} \phi_n(t)\}_{n \in \Gamma}$ as a collection of time derivatives of $\phi_n$ components.
\end{definition}
\begin{definition}
    For a given frame $\Phi=\{ \phi_n \}_{n \in \Gamma}$ consisting of functions of time, the zero-time frame operator is similar to the frame operator but only acts on $t=0$ instead of the integral over the entire domain:
\begin{equation}
    \mathbb{Q}_{\Phi} f = \vec{x} ,\quad x_n= f(t=0) \overline{\phi}_n(t=0) \, .
\end{equation}
\end{definition}
\begin{theorem} \label{eq:Theorem_translating_SSM}
    For the representation defined in Eq.~\ref{Scaling_representation}, the partial derivative of $\vec{c}$ with respect to $T$ is
\begin{equation}
    \frac{\partial}{\partial T} \vec{c} (T) = - \frac{1}{\theta} A \vec{c} (T)+ \frac{1}{\theta} B u(T), \quad A = \mathbb{U}_{\dot{\Phi}} \mathbb{U}^*_{\widetilde{\Phi}} + \mathbb{Q}_\Phi \mathbb{Q}^*_{\widetilde{\Phi}} 
    \label{eq:Translate_SSM}
\end{equation}
where $B$ is the complex conjugate of a vector containing members of the main frame evaluated at $T=1$ so we have $B=\{ \overline{\phi}_n (T=1)  \}_{n \in \Gamma}$. 
The $A$ operator can also be described using the matrix representation
\begin{equation}
    A_{i,j}= \overline{\phi}_i (0) \widetilde{\phi}_j (0)   + \int_0^1 \left[ \frac{\partial}{\partial t} 
 \overline{{\phi}}_{i}(t) \right]_{t=t'} \widetilde{\phi}_{j} (t') dt' \, . 
    \label{Translate_matrix}
\end{equation}
\end{theorem}

Proof is provided in Appendix~\ref{Proof_translating_SSM}. We will refer to the SSM with a scaling measure as translated-SaFARi.  

\subsection{SaFARi as generalization of HiPPO}

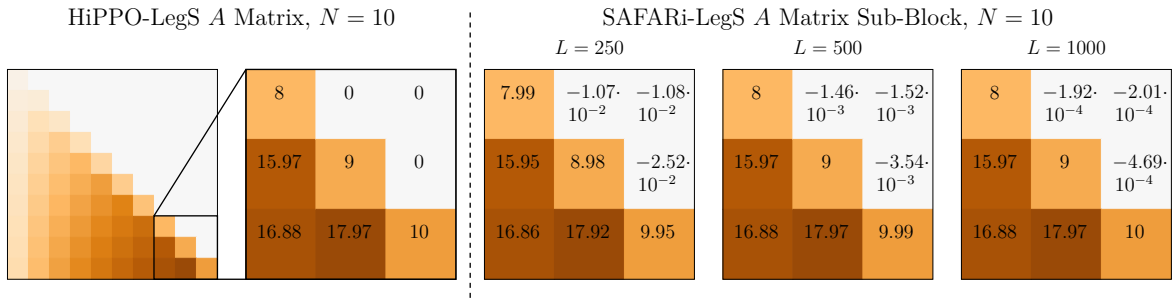
\begin{figure}[t!]
   \centering
        \resizebox{\textwidth}{!}{%}
        % \documentclass[tikz]{standalone}

% \usepackage{tikz}
% \usepackage{pgffor}
% \usetikzlibrary{fit}
% \usepackage{pgfplots}
% \usepackage{pgfplotstable}
% \usepackage{filecontents}
% \usepgfplotslibrary{groupplots}
% %\usetikzlibrary{pgfplots.groupplots}
% \pgfplotsset{compat=newest}
% \usepgfplotslibrary{colormaps}
% \usepgfplotslibrary{colorbrewer}

% \begin{document}

\pgfplotstableread[col sep=space, header=false] {Figures/safari_hippo/AHip.dat}\hippoA
\pgfplotstableread[col sep=space, header=false] {Figures/safari_hippo/Al_L250.dat}\Alow
\pgfplotstableread[col sep=space, header=false] {Figures/safari_hippo/Al_L500.dat}\Amid
\pgfplotstableread[col sep=space, header=false] {Figures/safari_hippo/Al_L1000.dat}\Ahigh

\begin{tikzpicture}
\begin{groupplot}[
    group style={
        group name=myplots,
        group size=5 by 1,
        horizontal sep=0.6cm,
    },
    width=6cm,
    height=6cm,
    /tikz/font=\large,
    tick style={draw=none},
    xtick=\empty, ytick=\empty,
    enlargelimits=true,
    axis on top,
    point meta min=-20,
    point meta max=20,
    enlargelimits=false,
    y dir=reverse,
    mesh/ordering=y varies,
    colormap/PuOr,
    ] 

\nextgroupplot[
    unbounded coords=jump,
    point meta = explicit,
    mesh/rows=10,
    mesh/cols=10,
    xmin=-0.5, xmax=9.5,
    ymin=-0.5, ymax=9.5,
    enlargelimits=false,
    ]
    \addplot [matrix plot*, point meta=explicit] 
    table [meta=2] {\hippoA};  
    
\nextgroupplot[
    point meta=explicit,
    mesh/ordering=rowwise, % Column-wise ordering of the matrix
    unbounded coords=jump,
    xmin=6.5,
    ymin=6.5,
    xmax=9.5,
    ymax=9.5,   
    enlargelimits=false,
    mesh/rows=3,
    ]
    \addplot [matrix plot*, point meta=explicit,
    mesh/cols=3, % Specify the number of columns
    mesh/rows=3, % Specify the number of rows
    mesh/check=false, % becaue it still complains about the number
    x filter/.expression={(x>=7 && y>=7) ? \pgfmathresult : NaN},
    unbounded coords=discard,
    ]
    table [x=0, y=1, meta=2] {\hippoA};
    
    \addplot [
        only marks, 
        point meta=explicit,
        nodes near coords={\pgfmathprintnumber\pgfplotspointmeta\,},
        x filter/.expression={(x<=y) ? \pgfmathresult : NaN},
        unbounded coords=jump,
        every node near coord/.append style={
            align=center,
        }
    ] table [meta=2] {\hippoA};  

     \addplot [
        only marks, 
        point meta=explicit,
        nodes near coords={\pgfmathprintnumber\pgfplotspointmeta\,},
        x filter/.expression={(x>y) ? \pgfmathresult : NaN},
        unbounded coords=jump,
        every node near coord/.append style={
            text width=1cm,
            align=center,
        }
    ] table [meta=2] {\hippoA}; 

\nextgroupplot[
    title={$L=250$},
    point meta=explicit,
    mesh/ordering=rowwise, % Column-wise ordering of the matrix
    unbounded coords=jump,
    xmin=6.5,
    ymin=6.5,
    xmax=9.5,
    ymax=9.5,   
    enlargelimits=false,
    mesh/rows=3,
    ]
    \addplot [matrix plot*, point meta=explicit,
    mesh/cols=3, % Specify the number of columns
    mesh/rows=3, % Specify the number of rows
    mesh/check=false, % becaue it still complains about the number
    x filter/.expression={(x>=7 && y>=7) ? \pgfmathresult : NaN},
    unbounded coords=discard,
    ]
    table [x=0, y=1, meta=2] {\Alow};
    
    \addplot [
        only marks, 
        point meta=explicit,
        nodes near coords={\pgfmathprintnumber\pgfplotspointmeta\,},
        x filter/.expression={(x<=y) ? \pgfmathresult : NaN},
        unbounded coords=jump,
        every node near coord/.append style={
            align=center,
        }
    ] table [meta=2] {\Alow};  

    \addplot [
        only marks, 
        point meta=explicit,
        nodes near coords={\pgfmathprintnumber\pgfplotspointmeta\,},
        x filter/.expression={(x>y) ? \pgfmathresult : NaN},
        unbounded coords=jump,
        every node near coord/.append style={
            text width=1cm,
            align=center,
            yshift = 15pt,
        }
    ] table [meta=2] {\Alow};   

\nextgroupplot[
    title={$L=500$},
    point meta=explicit,
    mesh/ordering=rowwise, % Column-wise ordering of the matrix
    unbounded coords=jump,
    xmin=6.5,
    ymin=6.5,
    xmax=9.5,
    ymax=9.5,   
    enlargelimits=false,
    mesh/rows=3,
    ]
    \addplot [matrix plot*, point meta=explicit,
    mesh/cols=3, % Specify the number of columns
    mesh/rows=3, % Specify the number of rows
    mesh/check=false, % becaue it still complains about the number
    x filter/.expression={(x>=7 && y>=7) ? \pgfmathresult : NaN},
    unbounded coords=discard,
    ]
    table [x=0, y=1, meta=2] {\Amid};
    
    \addplot [
        only marks, 
        point meta=explicit,
        nodes near coords={\pgfmathprintnumber\pgfplotspointmeta\,},
        x filter/.expression={(x<=y) ? \pgfmathresult : NaN},
        unbounded coords=jump,
        every node near coord/.append style={
            align=center,
        }
    ] table [meta=2] {\Amid};  

    \addplot [
        only marks, 
        point meta=explicit,
        nodes near coords={\pgfmathprintnumber\pgfplotspointmeta\,},
        x filter/.expression={(x>y) ? \pgfmathresult : NaN},
        unbounded coords=jump,
        every node near coord/.append style={
            text width=1cm,
            align=center,
            yshift = 15pt,
        }
    ] table [meta=2] {\Amid};   

\nextgroupplot[
    title={$L=1000$},
    point meta=explicit,
    mesh/ordering=rowwise, % Column-wise ordering of the matrix
    unbounded coords=jump,
    xmin=6.5,
    ymin=6.5,
    xmax=9.5,
    ymax=9.5,   
    enlargelimits=false,
    mesh/rows=3,
    ]
    \addplot [matrix plot*, point meta=explicit,
    mesh/cols=3, % Specify the number of columns
    mesh/rows=3, % Specify the number of rows
    mesh/check=false, % becaue it still complains about the number
    x filter/.expression={(x>=7 && y>=7) ? \pgfmathresult : NaN},
    unbounded coords=discard,
    ]
    table [x=0, y=1, meta=2] {\Ahigh};
    
    \addplot [
        only marks, 
        point meta=explicit,
        nodes near coords={\pgfmathprintnumber\pgfplotspointmeta\,},
        x filter/.expression={(x<=y) ? \pgfmathresult : NaN},
        unbounded coords=jump,
        every node near coord/.append style={
            align=center,
        }
    ] table [meta=2] {\Ahigh};  

    \addplot [
        only marks, 
        point meta=explicit,
        nodes near coords={\pgfmathprintnumber\pgfplotspointmeta\,},
        x filter/.expression={(x>y) ? \pgfmathresult : NaN},
        unbounded coords=jump,
        every node near coord/.append style={
            text width=1cm,
            align=center,
            yshift = 15pt,
        }
    ] table [meta=2] {\Ahigh};   

\end{groupplot}

\draw[thick,black] ([shift={(-38pt,38pt)}]myplots c1r1.south east) rectangle (myplots c1r1.south east);
\draw[thick,black]  (myplots c1r1.south east) -- (myplots c2r1.south west);
\draw[thick,black]  ([shift={(-38pt,38pt)}]myplots c1r1.south east) -- (myplots c2r1.north west);
\draw[thick,black]  (myplots c2r1.north west) rectangle (myplots c2r1.south east);

\node [fit={(myplots c1r1.south west)(myplots c2r1.north east)},inner sep=20pt,label={above:\Large{HiPPO-LegS $A$ Matrix, $N=10$}}] {};

\node [fit={(myplots c3r1.south west)(myplots c5r1.north east)},inner sep=20pt,label={above:\Large{SAFARi-LegS $A$ Matrix Sub-Block, $N=10$}}] {};

\draw[thick,black,dashed] ([shift={(.3cm,1.2cm)}]myplots c2r1.north east) -- ([shift={(-.3cm,-.5cm)}]myplots c3r1.south west);

\end{tikzpicture}

% \end{document}
    }
    \vspace{-0.8cm}
    \caption{HiPPO provides a closed-form solution for the scaled Legendre (LegS) SSM.  SaFARi provides a computed solution, where the accuracy depends on the discretization of the $N\times L$ frame.  Larger $L$ gives a finer discretization of the basis vectors and thus a better numerical result.}
    \label{fig:converge}
\end{figure}

HiPPO provides exact, closed-form solutions for $A$ and $B$ for a few specific basis and measure combinations. 
SaFARi replicates these $A$ and $B$ matrices to within some numerical error caused by discretization of the frame.
For a frame with $N$ components, the member vectors have a discretized length $L$.  
Increasing $L$ will provide matrices that converge toward the closed-form solution (see Fig. \ref{fig:converge}).
When a closed-form solution exists for the desired basis and measure (e.g., HiPPO-LegS for the scaled Legendre basis), then it is preferable to use it (see the derivations in \citet{gu2020hippo,gu2022train}).\footnote{Note that HiPPO used the convention of absorbing a negative sign from the ODE in Eq.~\ref{Time_Varying_SSM} into the $A$ matrix, where we do not.  
Additionally, authors in \citet{gu2020hippo,gu2022train} derived multiple versions for some $A$ matrices (e.g., FouT), and we find some minor discrepancies with our derivation.  See Appendix~\ref{sec:app_generalization} for details.}
SaFARi provides a method for \textit{any} basis/measure where the closed-form solution might not exist.
We also show that SaFARi preserves all of HiPPO's advantages in robustness to timescale when applied to a general frame in the Appendix~\ref{prop1_proof}

%\textbf{Time-scale robustness}: For scaled-SaFARi, dilating the input signal results in the same representation. Also, for the translated-SaFARi with parameter $\theta$, dilating the signal and the parameter $\theta$ at the same scale results in the same representation. The details are available in the Appendix~\ref{prop1_proof}.

\section{Implementation Considerations}\label{sec:implementation}
\vspace{-0.2cm}
This section describes the computational efficiency and accuracy concerns of SaFARi,
including strategies for producing the finite-dimensional approximation of the complete infinite-dimensional SaFARi in Section~\ref{sec:truncation}.
We analyze the errors introduced by these approximations in Section~\ref{sec:error_analysis}.

\subsection{Truncation of frames}\label{sec:truncation}
\vspace{-0.2cm}
Section~\ref{sec:SAFARI} demonstrates how a particular SSM can be built from an arbitrary frame $\Phi$.
Since the input space for the SSM is the class of functions of time, no $\Phi$ with a finite number of elements can meet the frame condition (Eq.~\ref{Frame_condition}), since the true representation of the input signal is infinite-dimensional.
In practice, the representation reduces to the truncated representation.
In this section, we analyze the theoretical implications of truncated representation using SaFARi.

\subsubsection{Finite-dimensional approximation of SaFARi}
\label{partial_reconstruction_section} \vspace{-0.2cm}
In the finite-dimensional case, we will use only $N$ elements of a frame.
Partial representation of size $N$ requires that the resulting representation approximates the infinite-dimensional representation.
We call the SSM with its $\vec{c}$ having $N$ coefficients $\textsf{SaFARi}^{(N)}$.
\begin{definition}
    \label{partial_WaLRUS-matrixes-def}
    A $\text{SaFARi}^{(N)}$ sequence is a sequence of the pairs $[ A^{(N)}, B^{(N)} ] $  where $ A^{(N)} \in \mathbb{C}^{N \times N} $ and $ B^{(N)} \in \mathbb{C}^{N} $ such that sequence converges to $[A,B]$ of SaFARi as
    \begin{equation}
        \lim_{N \rightarrow \infty} A^{(N)} = A, \qquad \lim_{N \rightarrow \infty} B^{(N)} = B \, .
    \end{equation} \vspace{-0.3cm}
 \end{definition}
It's trivial that this results in a finite-dimensional approximation of the infinite-dimensional representation.
Thus, any arbitrary precision can be achieved by selecting the appropriate truncation level.

Definition~\ref{partial_WaLRUS-matrixes-def} is not a constructive definition; that is, it does not uniquely determine $[ A^{(N)}, B^{(N)} ] $. 
In fact, there are many such sequences that all converge to $[A,B]$ of SaFARi. 
Of course, this does not mean that all such sequences produce equal representation error.  
We present and analyze two different constructions below. 

\subsubsection{Truncation of dual (ToD)} \vspace{-0.2cm}
The first construct of a $\text{SaFARi}^{(N)}$ sequence is through finding the infinite dimensional $A$ and $B$, then truncate to size $N$ 
\begin{equation}
\label{Truncatated_matrix}
    A^{(N)}= A_{[0:N , 0:N]}, \quad B^{(N)}= B_{[0:N ]} \, .
\end{equation} 
This construction results in a sequence that approximates the infinite-dimensional SaFARi according to Definition~\ref{partial_WaLRUS-matrixes-def}. 
The practical way of finding the truncated $A,B$ involves finding $A_{i,j}$ as introduced in Eq.~\ref{scaling_matrix} and Eq.~\ref{Translate_matrix} for $i,j < N$. 

Note that calculating $A_{i,j}$ requires finding $\widetilde{\Phi}$, the dual of the (infinite-dimensional) frame $\Phi$. 
For  certain families of functions, the dual frame can be found analytically.  
However, if an analytical way of finding $\widetilde{\Phi}$ is not known, then one must use a numerical approximation of the dual frame. 
In this case, the construction for $[ A^{(N)}, B^{(N)} ] $ involves forming the truncated frame for a much bigger size $N_2 \gg N$, then finding  $\widetilde{\Phi}^{(N_2)}= \rm{Dual} \{ \Phi_{[0:N_2]} \}$ numerically as an approximation for dual frame ($\widetilde{\Phi}^{(N_2)} \approx \widetilde{\Phi}$). 
Next, we truncate the approximate dual and use its first $N$ elements as an approximation for the first $N$ elements of the dual frame in constructions in Eq.~\ref{scaling_matrix} and Eq.~\ref{Translate_matrix}.

\subsubsection{Dual of truncation (DoT)}
\label{DoT_section}

The previous construction (ToD) becomes numerically intractable for cases where the dual frame is not analytically known; this motivates the need for an alternate constructor for \text{SaFARi}$^{(N)}$. To construct this sequence:
\begin{enumerate}
    \vspace{-0.2cm}
    \item Truncate the frame at level $N$ and form $\Phi^{(N)}= \{ \phi_i \}_{i<N}$.
    \item Numerically approximate the dual of the truncated frame $\widetilde{\Phi}^{(N)}= \rm{Dual} \{ \Phi^{(N)} \}$ using the pseudo-inverse.
    \item Use $\Phi^{(N)}$ and $\widetilde{\Phi}^{(N)}$ in Eq.~\ref{scaling_matrix} or Eq. \ref{Translate_matrix} to compute $[A^{(N)},B^{(N)} ]$.
\end{enumerate}

The constructed $[A^{(N)},B^{(N)} ]$ is a SaFARi$^{(N)}$ sequence according to Definition~\ref{partial_WaLRUS-matrixes-def}.

DoT and ToD approximate SaFARi with different rates. 
In the next sections, we provide a thorough error analysis of SaFARi$^{(N)}$ that enables us to compare the usage of different frames, as well as different construction methods for size $N$ constructs. 
We then demonstrate that the DoT construction always has a minimum reconstruction error, and is the optimal choice for implementing SaFARi.

\subsection{Error analysis} \label{sec:error_analysis}
Using truncated representations for online function approximation will result in some error in reconstruction, regardless of basis.
We focus here on errors emanating from this truncation of the representation in an SSM, rather than those caused by sampling, which have been extensively studied in the digital signal processing literature \citep{oppenheim1999discrete}.

Let $c^{(\infty)}$ denote the infinite-dimensional representation obtained from an SSM without truncating its associated $A,B$ matrices.
When the associated $A,B$ matrices for the SSM are truncated to the first N levels, it produces a truncated representation $c^{(\rm{N})}$. 
This truncation causes two distinct types of error, which we outline below.

\subsubsection{Truncation errors}
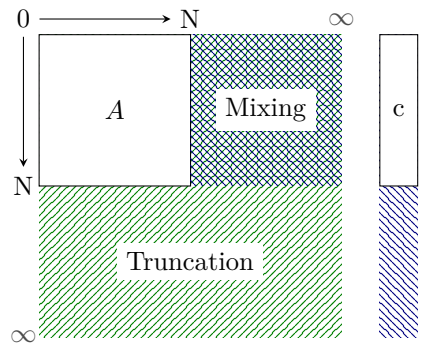
\begin{wrapfigure}{r}{0.35\textwidth}
\centering
        \resizebox{0.35\textwidth}{!}{%}
        % \documentclass[border=5mm]{standalone}
% \usepackage{tikz}
% \usetikzlibrary{calc}
% \usetikzlibrary{patterns, patterns.meta}
% \usepackage{xcolor}
% \colorlet{estColor}{green!40!black}
% \colorlet{errColor}{red}
% \colorlet{ubColor}{blue}
% \tikzset{>=latex} % for LaTeX arrow head
% \tikzstyle{vector}=[-stealth,thick,line cap=round]
% \usepackage{pgfplots}
% \usepackage{pgfplotstable}
% \pgfplotsset{compat=1.5}
% \usepgfplotslibrary{groupplots}

% \begin{document}

\begin{tikzpicture}
    \node [rectangle, 
    minimum width=4cm, minimum height=4cm,
    anchor = north west,
    pattern={Lines[angle=45,distance={2}]},pattern color=green!50!black]
    (one) at (0,0) {};
    \node [rectangle, draw, fill=white,
    anchor=north west,
    minimum width=2cm, minimum height=2cm] (A) at (0,0) {$A$};
    \node [rectangle, 
    minimum width=2cm, minimum height=2cm,
    anchor = north west,
    pattern={Lines[angle=-45,distance={2}]}, pattern color=blue!50!black] 
    (two) at (2cm,0) {};  
    \node [fill=white] (typeII) at (3cm,-1cm) {Mixing};
    \node [fill=white] (typeI) at (2cm,-3cm) {Truncation};

    \node [rectangle,
    minimum width=0.5cm, minimum height=4cm,
    anchor= north west,
    pattern={Lines[angle=-45,distance={2}]}, pattern color=blue!50!black] 
    (twoc) at (4.5cm,0) {};
    \node [rectangle,
    minimum width=0.5cm, minimum height=2cm,
    anchor=north west,
    draw, fill=white]
    (c) at (4.5cm,0) {c};

    \node (zero) at (-0.2cm,0.2cm) {0};
    \node (N1) at (-0.2cm,-2cm) {N};
    \node (N2) at (2cm,0.2cm) {N};
    \node [text=black!70] (infty1) at (-0.2cm,-4cm) {$\infty$};
    \node [text=black!70] (infty2) at (4cm,0.2cm) {$\infty$};

    \draw [-stealth] (zero) -- (N1);
    \draw [-stealth] (zero) -- (N2);
  
\end{tikzpicture}

% \end{document}
    }
    \caption{Illustration of error types in an SSM due to frame truncation. Truncation errors arise from signal energy in coefficients of index $n > N$, while mixing errors result from energy blending during $A \, c$.}
    \label{fig:errortypes}  
\end{wrapfigure}

Truncation errors are due to the fact that the truncated frame of size $N$ cannot represent any part of the signal contained in indices $n>N$.  
This is not limited to SSMs, but is true for any basis representation of a signal.
Intuitively, if we wish to represent a signal with, say, the first $N$ elements of a Fourier series, then any frequencies higher than $N$ will be lost.  
In SaFARi, truncation errors correspond to discarding the green shaded region in the $A^{(N)}$ illustrated in Fig.~\ref{fig:errortypes}. 

\subsubsection{Mixing errors} \vspace{-0.1cm}
Mixing errors arise from error propagation in the SSM update rule. 
Specifically, this update involves computing $A \,c$ (as in Eq.~\ref{Scaling_SSM} and Eq.~\ref{eq:Translate_SSM}), where the matrix $A$ introduces unwanted mixing between the omitted components ($n>N$) and the retained components ($n\leq N$) of the representation. Consequently, errors from the truncated portion of the representation propagate into the non-truncated portion. 
This is illustrated by the blue shaded regions in Fig. \ref{fig:errortypes}.
For the operator $A$ and truncation level $N$, the contaminating part of operator is $A_{i,j} \quad \forall (i\leq N , j>N)$. 

In the case of a translated measure, this mixing error is exacerbated since each update step requires estimating the initial value in that window (recall Eq.~\ref{eq:Theorem_translating_SSM}).
A key insight from our analysis below is that mixing errors have two sources:
\begin{enumerate}
    \vspace{-0.2cm}
    \item nonzero components in the upper right quadrant of $A$, and
    \vspace{-0.2cm}
    \item nonzero coefficients in $c$ at indices greater than $N$.
\end{enumerate} 

\subsubsection{Mitigating errors}

Truncation errors can never be eliminated, but may be alleviated by using a frame that exhibits a rapid decay in the energy carried by higher-order levels of representation.

\begin{figure}[t!]
    % \centering
    % \begin{minipage}[t]{0.65\textwidth}
        \centering
        \resizebox{\textwidth}{!}{%
            % \documentclass[tikz]{standalone}

% \usepackage{tikz}
% \usepackage{pgfplots}
% \usepackage{filecontents}
% \usetikzlibrary{pgfplots.groupplots}
% \pgfplotsset{compat=newest}

% \begin{document}
% \pgfplotsset{colormap={CM}{color=(black) rgb255=(113, 71, 71) rgb255=(184, 119, 119) 
% rgb255=(206, 167, 145) rgb255=(216, 193, 158) rgb255=(230, 226, 176) rgb255=(237, 237, 199) color=(white)}}
% \pgfplotsset{/pgfplots/colormap={CM}{rgb255=(0, 255, 0) rgb255=(0,200,200) rgb255=(0,0,255) rgb255=(89, 2, 250) rgb255=(0,0,0) rgb255=(233, 2, 250) rgb255=(255,0,0) rgb255=(250, 172, 2) rgb255=(255,255,0)}}
%\pgfplotsset{/pgfplots/colormap={CM}{rgb255=(255,0,0) rgb255=(255,100,100) rgb255=(0,0,0) rgb255=(100,100,255) rgb255=(0,0,255)}}
%\usepgfplotslibrary{colormaps}
\pgfplotsset{colormap/PuOr}
%\pgfplotsset{colormap name=PuOr}

\begin{tikzpicture}
\tikzstyle{every node}=[font=\large]
\begin{groupplot}[
    group style={
        group name=myplots,
        group size=4 by 1,
        ylabels at=edge left,
        %colormap/PuOr,
        %colormap name=PuOr,
        horizontal sep=0.5cm,
    },
    width=6cm,
    height=6cm,
    /tikz/font=\Large,
    tick style={draw=none},
    xtick=\empty, ytick=\empty,
    enlargelimits=false,
    axis on top,
    point meta min=-50,
    point meta max=50,
    enlargelimits=false,
    y dir=reverse,
    mesh/ordering=y varies,
] 
\nextgroupplot[title={\large{(a) HiPPO-LegS}}]
    \addplot [matrix plot*, point meta=explicit] table [meta index=2] {Figures/error-types/Als.dat};
\nextgroupplot[title={\large{(b) HiPPO-LegT}}]
    \addplot [matrix plot*, point meta=explicit] table [meta index=2] {Figures/error-types/Alt.dat};
\nextgroupplot[title={\large{(c) HiPPO-FouS}}]
    \addplot [matrix plot*, point meta=explicit] table [meta index=2] {Figures/error-types/Afs.dat};
\nextgroupplot[title={\large{(d) HiPPO-FouT}}]
    \addplot [matrix plot*, point meta=explicit] table [meta index=2] {Figures/error-types/Aft.dat};
\end{groupplot}
%stuff

\begin{axis}[name=axis2, anchor=west, 
    at={(myplots c4r1.east)},
    colorbar,     
    height=6cm,
    width=3cm,
    point meta min=-60,
    point meta max=60,
    xshift=-1cm,
    axis line style={draw=none},
    tick style={draw=none},
    xtick=\empty, ytick=\empty]
    \addplot [draw=none] coordinates {(0,0)};
\end{axis}

\node [rectangle,minimum height = 0.5cm, minimum width=0.5cm, anchor=north] at (axis2.south) (buffer) {};

\end{tikzpicture}

% \end{document}
        }
        \caption{Examples of the $A$ matrices of the HiPPO SSM for several basis/measure combinations: (a)~Scaled Legendre, (b)~Translated Legendre, (c)~Scaled Fourier, (d)~Translated Fourier. The dense non-zero elements in the upper right of (b) explain its poor performance compared to (a). The numerous small nonzero elements above the diagonal in (c) and (d) contribute to mixing errors over long sequences.}
        \label{fig:Amats}
    % \end{minipage}%
    % \hfill
    % \begin{minipage}[t]{0.32\textwidth}
    %     \centering
    %     \resizebox{\textwidth}{!}{%
    %         \input{Figures/error-types/error_type}
    %     }
    %     \caption{Illustration of error types in an SSM due to frame truncation. Truncation errors arise from signal energy in coefficients of index $n > N$, while mixing errors result from energy blending during $A \, c$.}
    %     \label{fig:errortypes}
    % \end{minipage}
\end{figure}
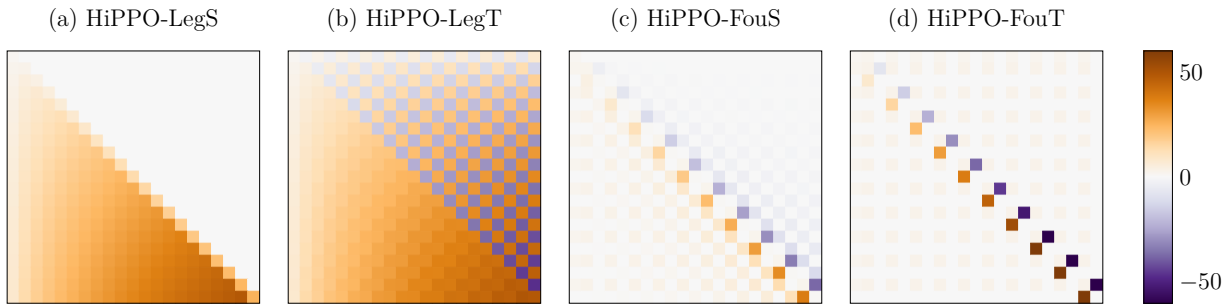

To counter mixing errors, we should ensure that values in the upper right quadrant of $A$ are as close to zero as possible, and/or  that coefficents of $c$ in the blue region of Fig.~\ref{fig:errortypes} are as close to zero as possible.  
If the matrix $A$ is lower-triangular, then any arbitrary truncation results in the contaminating part of $A$ being zero, which guarantees the second type of error is always zero, regardless of any coefficients in $c$.
This is the case for the HiPPO-LegS (scaled Legendre) $A$ matrix, as shown in Fig.~\ref{fig:Amats}(a). 
Indeed, the zero coefficents in the upper right quadrant of $A$ were considered strictly necessary in prior work \citep{gu2020hippo, gu2022train}.
This restriction explains the continued use of HiPPO-LegS in follow-up works, regardless of whether or not scaled Legendre polynomials were the typical choice for a given application.  

%\begin{figure}[t!]
    %\centering
    %\resizebox{\textwidth}{!}{%
    %    \input{Figures/error-types/A_mat}
    %}
%\caption{Examples of the $A$ matrices of the HiPPO SSM for several basis/measure combinations. (a)~Scaled Legendre, (b)~Translated Legendre, (c)~Scaled Fourier, (d)~Translated Fourier.  The density of non-zero elements in the upper right of (b) offers insight into its poor performance in contrast to (a).  The numerous small nonzero elements above the diagonal in (c) and (d) contribute to mixing errors over long sequences. 
%}
%\label{fig:Amats}
%\end{figure}

To summarize, there are two primary concerns when finding an appropriate frame for use in an SSM: 
\begin{enumerate}
    \vspace{-0.2cm}
    \item compatibility between the frame and the given class of input signals, and
    \vspace{-0.2cm}
    \item the operator $A$ that results from a given frame has a small contaminating part.
\end{enumerate}

Truncation and mixing errors have different sources but are linked. 
The optimal strategy to reduce both at the same time is to choose a basis that results in a representation where the energy is concentrated in the first $N$ coefficients. 
This can only be achieved, however, if we have some prior knowledge of the input signal in order to choose the right basis and truncation level.  
In cases where little is known about the input signal or correct truncation level, it is advisable to instead choose an SSM that is zero in the upper right quadrant, such as HiPPO-LegS, as it will inherently negate mixing errors.

\subsubsection{Error bound}

In order to quantify the mixing error,  
we show that the truncated representation follows the same differential equation with a perturbation defined by the theorem below.

\begin{theorem}
    \label{Theorem_perturbed_SSM} (Poof in Appendix~\ref{Proof_err_upperbound})
    The truncated representation generated by scaled-SaFARi follows a differential equation similar to the full representation, with the addition of a perturbation factor:
    \begin{equation}
    \frac{\partial}{ \partial T} c= - \frac{1}{T} A  c+ \frac{1}{T} B u(T) - \frac{1}{T} \vec{\epsilon}(T),
\end{equation}
where \vspace{-0.2cm}
\begin{equation}
    \vec{\epsilon}(T):= \langle  u_T , \xi \rangle, \qquad \xi = \Upsilon ( \widetilde{\Phi} \Phi - I ) \label{eq:def_xi}.
\end{equation}
\end{theorem}

The mixing error term $\vec{\epsilon}(T)$ cannot be directly calculated; however, we can derive an upper bound for the mixing error and demonstrate that this bound can be made arbitrarily small with appropriate truncation.

\begin{theorem}
\label{Theorem_err_upperbound}
If one finds an upper bound such that for all the times before $T$ we have $\| \epsilon(T)  \|_2 < \epsilon_m$, then the representation error is bound by
\begin{equation}
   \vspace{-0.2cm}
    \| \Delta c(T) \|_2 < \epsilon_m \sqrt{  \sum \frac{1}{\lambda^2_i} } = \epsilon_m \|  A^{-1} \|_F \, ; \vspace{-0.2cm}
\end{equation}
where $\lambda_i$ are the eigenvalues of $A$, and $F$ indicates the Frobenius norm. See proof in Appendix~\ref{Proof_err_upperbound}.
\end{theorem}

\subsubsection{Error analysis for different SaFARi\texorpdfstring{$\boldsymbol{^{(N)}}$}{} constructs}\label{sec:error_analysis_safariN}
In Section~\ref{partial_reconstruction_section} we define SaFARi$^{(N)}$ as a finite dimensional approximation for SaFARi, and provide two particular constructions, DoT and ToD. 
Armed with the quantification of representation error, we compare these different constructs. We provide Theorem~\ref{DoT_optimallity_theorem} to demonstrate DoT has the optimal representation error between different choices for SaFARi$^{(N)}$ using the same frame. 

\begin{theorem}
\label{DoT_optimallity_theorem}
    Given a frame $\Phi$, the Dual of Truncation (DoT) construct introduced in Section~\ref{DoT_section} has optimal representation error when compared to any other SaFARi$^{(N)}$ construct for the same frame $\Phi$. See proof in Appendix~\ref{DOT_optimallity_proof}.

\end{theorem}

As established by the result of Theorem~\ref{DoT_optimallity_theorem}, SaFARi$^{(N)}$ should be constructed with the DoT method.

\section{Computational efficiency and runtime complexity} \label{sec:computational_efficiency}
\vspace{-0.2cm}
This section develops the computational methods for obtaining sequence representations using SaFARi emphasizing its efficiency in both training and inference phases. 
We analyze the computational complexities of different update methods, highlight the role of parallel computation, and discuss the benefits of diagonalizable SSMs. 
These discussions provide a foundation for understanding the efficiency and scalability of SaFARi in practical applications.

\subsection{Computation complexity for sequential updates}
\vspace{-0.1cm}
To compute representations using scaled-SaFARi, the GBT update requires solving an $N \times N$ linear system at each step, leading to $O(N^3 L)$ complexity for a sequence of length $L$. In contrast, translated-SaFARi reuses the same inverse across all steps, reducing the complexity to $O(N^2 L)$.

If the state matrix $A$ is diagonalizable, both scaled and translated variants can be solved efficiently using a diagonalized SSM. The sequence representation $\widetilde{C}$ is computed for the diagonal SSM and transformed back via $C = V \widetilde{C}$, reducing the complexity to $O(N L)$. On parallel hardware, such diagonalized systems decompose into $N$ independent scalar SSMs, yielding $O(L)$ runtime with sufficient parallel resources.

If the state matrix $A$ is diagonalizable, both scaled and translated variants can accelerated via diagonalization. The sequence representation $\widetilde{C}$ is computed for the diagonal SSM and transformed back via $C = V \widetilde{C}$, reducing the complexity to $O(N L)$. On parallel hardware, such diagonalized systems decompose into $N$ independent scalar SSMs, yielding $O(L)$ runtime with sufficient parallel resources.

Diagonalizability of an SSM depends on the frame or basis used in its construction. One limitation of Legendre-based SSMs such as HiPPO is that its $A$ matrix cannot be stably diagonalized, even for representation sizes as small as $N=20$ \citep{diag_init_2022}, leading to significantly higher cost.

To address this limitation, \citep{gu2022train} proposed a fast sequential update method for HiPPO-LegS, claiming $O(N)$ computational complexity and $O(1)$ runtime complexity per update on parallel processors.
However, we observe that this method becomes numerically unstable at larger $N$, as discussed in Appendix~\ref{section:fast_legs_failure}
To resolve this, we suggest a simple modification: computing lower-degree representations before higher-degree ones. 
While the modified approach perserves the $O(N)$ computational cost,  it increases runtime to $O(N)$ per sequential update, as it is no longer parallelizable.

Similar to HiPPO-LegS for the scaled measure, HiPPO-LegT for the translated measure also cannot be stably diagonalized.  
To our knowledge, LegT has no analogue for the fast LegS update method in \citet{gu2020hippo}.
Therefore, the computational complexity remains considerably higher than for diagonalizable SSMs.

\subsection{Convolution kernels and diagonalization}  \label{Sec:convolution_kernels} \vspace{-0.2cm}
When using an SSM for a recognition or learning task, a training phase is required in which the downstream model is trained on the generated representation. 
Using sequential updates for training is prohibitively taxing on computational and time resources as the whole sequence is available in the training time.
%A much faster method might be to simply form sub-sequences of the training data starting from the beginning of the sequence to every point in the sequence, and then project these "online history" of the training data onto the frame directly, but this approach will require excessive storage and computing resources for long data sequence.
Ideally, we would perform computations of the sequential SSM in parallel.   
However, this is a challenge since each new update depends on the result of the previous.  
The authors of \citet{gu2022s4} discussed how to implement a parallel computation algorithm for SSMs produced by HiPPO. 
To do so, one ``unrolls'' the SSM as:
\vspace{-0.1cm}
\begin{equation}
c_k= \overline{A}_k \dots \overline{A}_1  \overline{B}_0 u_0 + \dots + \overline{A}_k \overline{B}_{k-1} u_{k-1} + \overline{B}_k u_k \quad , \quad c= K * u
\label{eq:SSM_unrolled}
\end{equation}
\begin{equation}
    \overline{A}_i = (  I + \delta t \alpha A_i )^{-1} (  I - \delta t (1-\alpha )  A_i), \qquad \overline{B}_i = (  I + \delta t \alpha A_{i+1} )^{-1} B_i.
\end{equation}
The convolution kernel \( K \) in Eq.~\ref{eq:SSM_unrolled} removes the sequential dependency, enabling parallel computation on hardware such as GPUs, and significantly reducing training time.

\subsection{Runtime complexity of scaled-SaFARi} \label{Sec:complexity_analisys_scaled}
\vspace{-0.2cm}
For scaled-SaFARi, if the SSM is not diagonalizable, then the kernel can still be computed in parallel by framing the problem as a scan over the associative prefix product. Since matrix multiplication is associative, all such prefix products can be computed efficiently using parallel scan algorithms \citep{blelloch1990prefix, hillis1986data}. When implemented on parallel hardware such as GPUs, this strategy achieves a time complexity of $O ( N^3 logL)$ if enough parallel processors are available.

However, if the SSM is diagonalizable, all the matrix products $\bar{A}_k$ matrices in the matrix products appearing in the kernel expression become diagonal. As a result, the convolution kernel can be calculated with the time complexity of $O ( N logL)$. Furthermore, the below theorem suggests how to find the kernel in closed form.

\begin{theorem} \label{Theorem_scaling_kernel}
For scaled-SaFARi, if $A$ is diagonalizable, then the convolution kernel $K$ that computes the representations can be found in closed form. See the appendix for the closed form formula. See Appendix~\ref{Proof_scaling_kernel}.
\end{theorem}

As noted above, HiPPO-LegS is not diagonalizable, complicating kernel computation. A heuristic method proposed by \citet{gu2022train} enables approximate kernel evaluation with time complexity $O(N \log L)$, but it introduces additional computation and lacks a closed-form solution.

\subsection{Runtime complexity of translated-SaFARi} \label{Sec:complexity_analisys_translated}
\vspace{-0.2cm}
Similar to the scaled case, all the matrix products $\bar{A}_k \dots \bar{A}_{k-m}$ can be computed efficiently using parallel scan algorithms \citep{blelloch1990prefix, hillis1986data}. As a result, the convolution kernel $K$ can be computed with an overall time complexity of $O( N^2 log(L) )$ since $\bar{A}_k$ remains the same for all the values of $k$. Similar to the scaled case, if the $A$ matrix is diagonalizable, then the $\bar{A}$ matrices becomes diagonal.
This lowers down the time complexity of finding. With access to enough parallel processors, the runtime complexity can be further reduced to $O( N^2 log(L) )$.

Furthermore, for diagonalizable SSMs, the convolution kernel for the translated measure has a closed form. 

\begin{theorem} \label{Theorem_translating_kernel}
For translated-SaFARi, if $A$ is diagonalizable, then the convolution kernel $K$ that computes the representations can be found in closed form. See the appendix for the closed form formula. See Appendix~\ref{Proof_translating_kernel}
\end{theorem}

\section{Conclusions}\label{sec:conclusions}
\vspace{-0.3cm}
In this work, we demonstrate how function approximation with SSMs, originally introduced by \citet{gu2020hippo}, can be extended to general frames through SaFARi. 
Our method generalizes the HiPPO framework to accommodate bases or frames where closed-form solutions are unavailable, paving the way for broader applicability in sequential modeling tasks. 
We provide guidelines for frame selection and SSM construction in Sec.~\ref{sec:implementation}, highlighting the importance of both the representational power of the chosen frame and the structural properties of the $A$ matrix in the SSM.

The versatility of SaFARi unlocks new opportunities for exploring novel SSM-based architectures in machine learning and signal processing. Future directions include the evaluation of wavelet and other structured frames within SaFARi, aiming to enhance localized and sparse representations beyond the capabilities of polynomial or Fourier methods. Moreover, integrating SaFARi into advanced SSM architectures like S4 and Mamba offers promising avenues for improving long-range sequence modeling.

By extending SSM construction beyond specific bases, SaFARi provides a flexible foundation for adaptive, efficient state-space representations, linking theoretical advancements to practical applications in sequential data processing.

\section{Acknowledgments}
Special thanks to T. Mitchell Roddenberry for fruitful conversations over the course of this project. 
This work was supported by NSF grants CCF-1911094, IIS-1838177, and IIS-1730574; ONR grants N00014-18-12571, N00014-20-1-2534, N00014-18-1-2047, and MURI N00014-20-1-2787; AFOSR grant FA9550-22-1-0060; DOE grant DE-SC0020345; and a Vannevar Bush Faculty Fellowship, ONR grant N00014-18-1-2047, Rice Academy of Fellows, and Rice University and Houston Methodist 2024 Seed Grant Program.

\bibliography{ref}
\bibliographystyle{tmlr}

\appendix
\section{Appendix}
\renewcommand{\theequation}{\thesection.\arabic{equation}}
\setcounter{equation}{0}

\subsection{Derivation of SaFARi with the scaled measure}
\begin{apptheorem}[\ref{Theorem_Scaling_SSM}]
\label{Proof_scaling}
  Assuming elements in the frame $\Phi$ and the input signal $u$ are right-continuous, for the representation defined in Eq.~\ref{Scaling_representation},
the partial derivative is:
\begin{equation}
    \frac{\partial}{\partial T} \vec{c} (T) = - \frac{1}{T} A \vec{c} (T)+ \frac{1}{T} B u(T)
\end{equation}
where $A$ is a linear operator defined as:
\begin{equation}
    A= I+ \mathbb{U}_{\Upsilon}  \mathbb{U}^*_{\widetilde{\Phi}} 
\end{equation}
and $B$ is the complex conjugate of a vector containing members of the main frame evaluated at $T=1$ so we have $B= \{ \overline{\phi}_n (T=1) \}_{n \in \Gamma}$ . One can show that the  $A$ operator can also be described using the matrix multiplication:
\begin{equation}
    A_{i,j}=\delta_{i,j} + \int_0^1 \overline{\upsilon}_{i}(t) \widetilde{\phi}_{j} (t) dt 
    %\label{scaling_matrix}
\end{equation}
\end{apptheorem}
\textbf{Proof}: 
Taking partial derivative with respect to $T$, we have:
\begin{equation}
    \frac{\partial }{\partial T}c_n(T)= \frac{\partial}{\partial T}\int_{t=0}^T \frac{1}{T} u(t) \overline{  \phi_n } \left(\frac{t}{T}\right)  dt
\end{equation}
Now applying the Leibniz integral rule we have:
\begin{align}
    \frac{\partial }{\partial T}c_n(T)&= \int_{t=0}^T \frac{-1}{T^2} u(t) \overline{  \phi_n } \left(\frac{t}{T}\right)  dt + \int_{t=0}^T  \frac{-1}{T^2}  u(t) \frac{t}{T}\overline{  \phi'_n } (\frac{t}{T})  dt + \frac{\overline{\phi}_n(1)}{T} u(T) \nonumber \\
    &= \frac{-1}{T} c_n + \frac{-1}{T} \int_{t=0}^T u(t) 
  \frac{1}{T} \overline{  \upsilon }_n \left(\frac{t}{T}\right)  dt + \frac{\overline{\phi}_n(1)}{T} u(T) \nonumber \\
  & = \frac{-1}{T} c_n + \frac{-1}{T} \langle u , \upsilon_n  \rangle + \frac{\overline{\phi}_n(1)}{T} u(T)
    \label{Deriv1}
\end{align}

This is still not an SSM since the second term is not explicitly a linear form of $\vec{c}(T)$. To convert this to a linear form of $\vec{c}(T)$, we use the equality given in Eq.~\ref{Synthesis} to represent $\upsilon_n$ using the frame $\widetilde{\Phi}$
\begin{equation}
    \upsilon_n(t)= \sum_{j \in \Gamma} \langle \upsilon_n , \phi_j    \rangle \widetilde{\phi}_j=\sum_{j \in \Gamma} \langle \upsilon_n , \widetilde{\phi}_j    \rangle \phi_j
    \label{Representing_Nu}
\end{equation}
\begin{equation}
    \rightarrow \langle u , \upsilon_n  \rangle= \langle u,
    \sum_{j \in \Gamma} \langle \upsilon_n , \widetilde{\phi}_j    \rangle \phi_j \rangle =  \sum_{j \in \Gamma} \overline{ \langle \upsilon_n , \widetilde{\phi}_j    \rangle  }\langle u,
      \phi_j \rangle =\sum_{j \in \Gamma} \overline{ \langle \upsilon_n , \widetilde{\phi}_j \rangle } \,    c_j(T)
      \label{Upsilon_expansion_theorem1}
\end{equation}
Putting Eq.~\ref{Upsilon_expansion_theorem1} in Eq.~\ref{Deriv1} results in:
\begin{equation}
    \frac{\partial }{\partial T} \vec{c}(T)= - \frac{1}{T} (I+ \mathbb{U}_{\Upsilon}  \mathbb{U}^*_{\widetilde{\Phi}}   ) \vec{c} (T) + \frac{\overline{\phi}_n(1)}{T} u(T) 
\end{equation}
This proves the theorem. $\square$

\subsection{Derivation of SaFARi with the translated measure}
\begin{apptheorem}[\ref{eq:Theorem_translating_SSM}] \label{Proof_translating_SSM}
    Assuming elements in the frame $\Phi$ and the input signal $u$ are right-continuous, for the representation defined in Eq.~\ref{Scaling_representation}, the partial derivative is:
\begin{equation}
    \frac{\partial}{\partial T} \vec{c} (T) = - \frac{1}{\theta} A \vec{c} (T)+ \frac{1}{\theta} B u(T)
\end{equation}
where $A$ is a linear operator defined as:
\begin{equation}
    \overline{A}= \mathbb{U}_{\dot{\Phi}} \mathbb{U}^*_{\widetilde{\Phi}} + \mathbb{Q}_\Phi \mathbb{Q}^*_{\widetilde{\Phi}}
\end{equation}
and $B$ is the complex conjugate of a vector containing members of the main frame evaluated at $T=1$ so we have $B=\{ \overline{\phi}_n (T=1)  \}_{n \in \Gamma}$. One can show that the  $A$ operator can also be described using the matrix multiplication:
\begin{equation}
    A_{i,j}= \overline{\phi}_i (0) \widetilde{\phi}_j (0)   + \int_0^1 \left[ \frac{\partial}{\partial t} 
 \overline{{\phi}}_{i}(t) \right]_{t=t'} \widetilde{\phi}_{j} (t') dt' 
    %\label{Translate_matrix}
\end{equation}
\end{apptheorem}

\textbf{Proof}: We can write the partial derivative as:
\begin{equation}
    c_n(T)= \int_{T-\theta}^T u(t) \frac{1}{\theta} \overline{\phi}_n\left( \frac{t - (T-\theta)}{\theta} \right)  dt
\end{equation}
Taking the partial derivative with respect to $T$, we have
\begin{equation}
    \frac{\partial c_n(T)}{ \partial T} = \frac{-1}{\theta^2} \int_{T-\theta}^T u(t) \overline{\phi}_n'\left( \frac{t - (T-\theta)}{\theta} \right)    dt
    +\frac{1}{\theta} \overline{\phi}_n(1) u(T) - \frac{1}{\theta} \overline{\phi}_n(0) u(T-\theta) 
    \label{Thorem1_eq1}
\end{equation}
Similar to our approach for the previous theorem, we write $ \phi'_n(z)$ as
\begin{equation}
     \phi'_n(z)= \sum_{i \in \Gamma} \langle    \phi'_n  ,\widetilde{\phi}_i  \rangle \phi_i(z) = \sum_{i \in \Gamma} Q_{n,i} \, \phi_i(z) 
\end{equation}
\begin{equation}
    Q_{n,i} = \int_{z=0}^{1} \phi'_n(z) \overline{\widetilde{\phi}}_i(z) dz 
\end{equation}
Now, if we use this expansion, and put it in Eq.~\ref{Thorem1_eq1} we have
\begin{equation}
    \frac{\partial c_n(T)}{ \partial T} = \frac{-1}{\theta^2} \int_{T-\theta}^T  u(t) \left[ \sum_{i \in \Gamma} \overline{Q}_{n,i} \, \overline{\phi}_i\left( \frac{t - (T-\theta)}{\theta}  \right)  \right] dt
    +\frac{1}{\theta} \overline{\phi}_n(1) u(T) - \frac{1}{\theta} \overline{\phi}_n(0) u(T-\theta) 
\end{equation}
\begin{equation}
    \frac{\partial c_n(T)}{ \partial T} = \frac{-1}{\theta} \sum_{i \in \Gamma} \overline{Q}_{n,i} \, \left[ \frac{-1}{\theta} \int_{T-\theta}^T  u(t)   \overline{\phi}_i\left( \frac{t - (T-\theta)}{\theta}  \right)   dt \right]
    +\frac{1}{\theta} \overline{\phi}_n(1) u(T) - \frac{1}{\theta} \overline{\phi}_n(0) u(T-\theta) 
\end{equation}
We also write $u(T-\theta)$ as a reconstruction using the current representation $ u(T-\theta)= \sum_i c_i \widetilde{\phi}_i(0) $
\begin{equation}
    \frac{\partial c_n(T)}{ \partial T} = \frac{-1}{\theta} \sum_{i} \overline{Q}_{n,i} \, c_i
    +\frac{1}{\theta} \overline{\phi}_n(1) u(T) - \frac{1}{\theta} \overline{\phi}_n(0) \sum_i c_i \widetilde{\phi}_i(0)
\end{equation}

\begin{equation}
    \frac{\partial c_n(T)}{ \partial T} = \frac{-1}{\theta} \sum_i (  \overline{Q}_{n,i} + \widetilde{\phi}_i(0) \overline{\phi}_n(0) )   \, c_i  +\frac{1}{\theta} \overline{\phi}_n(1) u(T) 
\end{equation}

If we put $A_{i,j}= \overline{Q}_{i,j} + \overline{\phi}_i(0) \widetilde{\phi}_j(0)$, it proves the theorem. $\square$

\subsection{Comparison of SaFARi and HiPPO $A$ matrices}
\label{sec:app_generalization}

Our results with SaFARi converge to HiPPO for LegS, and for the version of LegT given in \citet{gu2022train}.
We noted that there were some discrepancies with published versions of FouT.  
In \citet{gu2022train}, a complete closed-form solution is given as follows (flipping negative signs to be consistent with our notation):
\begin{equation}
A_{n,k}=
\begin{cases}
2 & n=k=0 \\
2\sqrt{2} & n=0, k \text{ odd} \\
2\sqrt{2} & k=0, n \text{ odd} \\
4 & n,k \text{ odd} \\
-2\pi k & n-k=1, k \text{ odd} \\
2\pi n & k-n=1, n \text{ odd} \\
0 & \text{otherwise}
\end{cases}
\end{equation}
Our derivation for a closed-form solution is different by a factor of 2, and has different values on the off-diagonals:
\begin{equation}
A_{n,k}=
\begin{cases}
1 & n=k=0 \\
\sqrt{2} & n=0, k \text{ odd} \\
\sqrt{2} & k=0, n \text{ odd} \\
2 & n,k \text{ odd} \\
\pi (k+1) & n-k=1, k \text{ odd} \\
-\pi (n+1) & k-n=1, n \text{ odd} \\
0 & \text{otherwise}
\end{cases}
\end{equation}

The discrepancy may be due to \citeauthor{gu2022train} using a Fourier basis other than the DFT over the range $[0,1]$ normalized to unit energy. 

\subsection{Mathematical properties of SaFARi}
\begin{appprop} \label{prop1_proof}
    For any scalar $\beta>0$, if $h(t)= u(\beta t)$ then for the scaled measure we have $\rm{SaFARi} (h) (T)=\rm{SaFARi} (u) (\beta T) $
\end{appprop}

\textbf{Proof}: We start by writing the representation generated by the scaled SaFARi for $h(t)$.
\begin{align}
    \rm{SaFARi} (h) (T) &= \int_{t=0}^T h(t) \overline{  \phi_n } \left( \frac{t}{T}\right) \frac{1}{T} dt \\
    &= \int_{t=0}^T u(\beta t) \overline{  \phi_n } \left(\frac{t}{T}\right) \frac{1}{T} dt
\end{align}
\begin{equation}
     \int_{t=0}^T u(t') \overline{  \phi_n } \left(\frac{t'}{\beta T}\right) \frac{1}{T} \frac{dt'}{\beta} = \rm{SaFARi} (u) (\beta T)
\end{equation}

\begin{appprop} \label{prop2_proof}
    For any scalar $\beta>0$, if $h(t)= u(\beta t)$ then for the translated measure with parameter $\theta$ we have $\rm{SaFARi}_\theta (h) (T)=\rm{SaFARi}_{\beta \theta} (u) (\beta T) $
\end{appprop}
Similar to the previous proposition, we start by writing the representation of the scaled SaFARi:
\begin{align}
    \rm{SaFARi}_\theta (h) (T) &= \int_{t=T-\theta}^T h(t) \overline{  \phi_n } \left( \frac{t-(T-\theta)}{\theta}\right) \frac{1}{\theta} dt \\
    &= \int_{t=T-\theta}^T u(\beta t) \overline{  \phi_n } \left( \frac{t-(T-\theta)}{\theta}\right) \frac{1}{\theta} dt \\
    &= \int_{t'=\beta T- \beta \theta}^{\beta T} u(t') \overline{  \phi_n } \left( \frac{t'-(\beta T-\beta \theta)}{\beta \theta}\right) \frac{1}{\theta} \frac{dt'}{\beta} = \rm{SaFARi}_{\beta \theta} (u) (\beta T) 
\end{align}

\subsection{The closed-form solution for SaFARi differential equations}

\begin{lemma}
    \label{Lemma1}
    The closed form solution for the differential equation introduced in Eq.~\ref{Scaling_SSM} is:

\begin{equation}
    c =  \int_{t=0}^T  \frac{1}{t} \exp{\left( A \ln \frac{t}{T} \right)}  B u(t) dt 
\label{eq:Solving_ssm_scaling}.
\end{equation}

\end{lemma}

\textbf{Proof}: We begin by re-writing the differential equation for any time $t$
\begin{equation}
    \frac{\partial}{\partial t} \vec{c} (t) + \frac{1}{t} A \vec{c} (t)= \frac{1}{t} B u(t)
\end{equation}
Now we multiply both sides by $  \exp(A \ln(t))$
\begin{equation}
    \exp(A \ln(t)) \frac{\partial}{\partial t} \vec{c} (t) + \frac{1}{t} A \exp \left(A \ln(t)\right) \vec{c} (t)= \frac{1}{t} \exp(A \ln(t)) B u(t)
\end{equation}
The left side of the equality is now a complete differential
\begin{equation}
    \frac{\partial}{\partial t} \left( \exp(A \ln(t))  \vec{c}(t) \right) = \frac{1}{t} \exp \left(A \ln(t)\right) B u(t)
\end{equation}
\begin{equation}
     \exp\left(A \ln(T)\right)  \vec{c}(T)  =\int_{t=0}^T \frac{1}{t} \exp\left(A \ln(t)\right) B u(t)
\end{equation}
\begin{equation}
       \vec{c}(T)  = \exp\left(-A \ln(T)\right) \int_{t=0}^T \frac{1}{t} \exp \left( A \ln(t) \right) B u(t)
\end{equation}
This proves Lemma 1. $\square$

\begin{lemma}
    \label{Lemma2}
    The closed form solution for the differential equation introduced in Eq.~\ref{eq:Translate_SSM} is:
\begin{equation}
    c = \int_{t=T-\theta}^T \frac{1}{\theta} \exp \left(A \frac{t-T}{\theta}\right) B u(t).
\label{eq:Solving_ssm_translating}
\end{equation}

\end{lemma}

\textbf{Proof}: We begin by re-writing the differential equation for any time $t$ as
\begin{equation}
    \frac{\partial}{\partial t} \vec{c} (t) + \frac{1}{\theta} A \vec{c} (t)= \frac{1}{\theta} B u(t).
\end{equation}
Now we multiply both sides by $  \exp(A \frac{t}{\theta})$
\begin{equation}
    \exp \left(A \frac{t}{\theta}\right) \frac{\partial}{\partial t} \vec{c} (t) + \frac{1}{\theta} A \exp \left(A \frac{t}{\theta}\right) \vec{c} (t)= \frac{1}{\theta} \exp \left(A \frac{t}{\theta}\right) B u(t)
\end{equation}
The left side of the equality is now a complete differential
\begin{equation}
    \frac{\partial}{\partial t} \left( \exp \left(A \frac{t}{\theta}\right)  \vec{c}(t) \right) = \frac{1}{\theta} \exp \left(A \frac{t}{\theta} \right) B u(t)
\end{equation}
\begin{equation}
       \vec{c}(T)  = \exp \left(-A \frac{T}{\theta}\right) \int_{t=T-\theta}^T \frac{1}{\theta} \exp \left(A \frac{t}{\theta} \right) B u(t)
\end{equation}
This proves Lemma 2. $\square$ 

\subsection{Error analysis}
\begin{apptheorem}[\ref{Theorem_perturbed_SSM}] \label{Proof_perturbed_SSM}
    The truncated representation generated by the scaled-SaFARi follows a differential equation similar to the full representation, with the addition of a perturbation factor.
    \begin{equation}
    \frac{\partial}{ \partial T} c= - \frac{1}{T} A  c+ \frac{1}{T} B u(T) - \frac{1}{T} \vec{\epsilon}(T).
    \end{equation}
where $\vec{\epsilon}$ is defined as
\begin{equation}
    \vec{\epsilon}(T)= \langle  u_T , \xi \rangle
\end{equation}
\begin{equation}
    \xi = \Upsilon ( \widetilde{\Phi} \Phi - I )
\end{equation}

\end{apptheorem}

\textbf{Proof}: Repeat the steps taken in the proof of Theorem~\ref{Theorem_Scaling_SSM} until Eq.~\ref{Representing_Nu}. Truncating the frame results in an error in this step which can be written as
\begin{equation}
    \upsilon_n(t)= \sum_{j \in \Gamma} \langle \upsilon_n , \widetilde{\phi}_j    \rangle \phi_j + \xi_n(t)
\end{equation}
In fact, this is how $\xi$ is defined. 
Adding $\xi$ as a correction term here changes the SSM derivation:
\begin{equation}
    \rightarrow \langle u , \upsilon_n  \rangle= \langle u,
    \xi_n+ \sum_{j \in \Gamma} \langle \upsilon_n , \widetilde{\phi}_j    \rangle \phi_j \rangle  =  \sum_{j \in \Gamma} \overline{ \langle \upsilon_n , \widetilde{\phi}_j    \rangle  }\langle u,
      \phi_j \rangle + \langle u, \xi_n \rangle =\sum_{j \in \Gamma} \overline{ \langle \upsilon_n , \widetilde{\phi}_j \rangle } \,    c_j(T) + \langle u, \xi_n \rangle
      \label{Upsilon_expansion1}
\end{equation}
Putting Eq.~\ref{Upsilon_expansion1} in Eq.~\ref{Deriv1} results in:
\begin{equation}
    \frac{\partial}{ \partial T} c= - \frac{1}{T} A  c+ \frac{1}{T} B u(T) - \frac{1}{T} \vec{\epsilon}(T). \qquad \square 
\end{equation}

\begin{apptheorem}[\ref{Theorem_err_upperbound}] 
    If one finds an upper bound such that  $\forall t<T$ we have $\| \epsilon(t)  \|_2 < \epsilon_m$, then the representation error can be bound by
\begin{equation}
    \| \Delta c(T) \|_2 < \epsilon_m \sqrt{  \sum \frac{1}{\lambda^2_i} } = \epsilon_m \|  A^{-1} \|_F
\end{equation}
\end{apptheorem}

\textbf{Proof}:\label{Proof_err_upperbound}  Using the result of Theorem~\ref{Theorem_perturbed_SSM}:
\begin{equation}
    \frac{\partial}{ \partial T} c= - \frac{1}{T} A  c+ \frac{1}{T} \left(B u(T) -\vec{\epsilon}(T) \right)
\end{equation}
We can use Lemma~\ref{Lemma1} to find the closed form solution of the perturbed SSM above
\begin{equation}
c = \int_{t=0}^{T} \frac{1}{t} \exp \left(A \ln{\frac{t}{T}}\right) \left( B u(t) -\vec{\epsilon}(t)\right) dt   
\end{equation}
\begin{equation}
c = \int_{t=0}^{T} \frac{1}{t} \exp \left(A \ln \frac{t}{T}\right) B u(t)  dt - \int_{t=0}^{T} \frac{1}{t} \exp\left(A \ln \frac{t}{T}\right) \vec{\epsilon}(t)  dt   
\end{equation}
\begin{equation}
\rm{Size \, N \, representation} = \rm{True \, representation} -  \rm{Error}
\end{equation}
The last term is indeed the second type error that we have discussed in the error analysis section of the paper. Using eigenvalue decomposition of $A = V \Lambda V^{-1}$ we re-write the above error term as
\begin{equation}
 \rm{Error}= V \int_{t=0}^{T} \frac{1}{t} \exp \left(\Lambda \ln \frac{t}{T}\right) 
 V^{-1} \vec{\epsilon}(t)  dt   
\end{equation}
with a change of variable $s= \ln \frac{t}{T}$
\begin{equation}
 V^{-1} \rm{Error}=  \int_{t=0}^{T} \exp(\Lambda s) 
 V^{-1} \vec{\epsilon}(s)  ds   
\end{equation}
According to the assumption of this theorem $\|V^{-1} \epsilon(t)  \|_2 = \| \epsilon(t)  \|_2 \leq \epsilon_m  $
\begin{equation}
 \rightarrow [V^{-1} \rm{Error}]_j \leq \int_{t=0}^{T} \exp(s \lambda_j) \epsilon_m ds  = \frac{\epsilon_m}{\lambda_j}
\end{equation}
\begin{equation}
    \| Error\|^2=\| V^{-1} Error\|^2 \leq \epsilon_m  \sqrt{  \sum \frac{1}{\lambda^2_i} } = \epsilon_m \|  A^{-1} \|_F 
\end{equation} 

\begin{apptheorem}[\ref{DoT_optimallity_theorem}]
    Suppose a frame $\Phi$ is given. The Dual of Truncation (DoT) construct introduced in Section~\ref{DoT_section} has optimal representation error when compared to any other SaFARi$^{(N)}$ construct for the same frame $\Phi$.
\end{apptheorem}

\textbf{Proof}:\label{DOT_optimallity_proof} The proof for this theorem involves two steps.

\begin{enumerate}
    \item First, we show that the optimal representation error in the theorem can be reduced to optimal error of the second type (mixing). 
    \item Then, we demonstrate that for a given frame $\Phi$, and given truncation level $N$, the construct with the optimal second type error control is DoT.
\end{enumerate}

As discussed in Sec.~\ref{sec:error_analysis}, the first type of error is due to truncating the frame, and is independent of the SSM. 
In the scope of this theorem, all the SaFARi$^{(N)}$ constructs use the same frame and the same truncation. Therefore, comparing the representation error between SaFARi$^{(N)}$ in the theorem reduces to comparing the mixing error.

The mixing error is shown in Theorem~\ref{Theorem_perturbed_SSM} to be proportional to
\begin{equation}
    \vec{\epsilon}(T)= \langle  u_T , \xi \rangle
\end{equation}
To minimize $\| \vec{\epsilon}(T) \|$ irrespective of the input signal, one has to minimize $\| \xi \|^2_F$

\begin{equation}
    \xi = \Upsilon_{[0:N]} ( \widetilde{\Phi}^{(N)} \Phi_{[0:N]} - I )
    \label{Xi_partial}
\end{equation}

Where $\Upsilon_{[0:N]}$ and $\Phi_{[0:N]}$ are the first $N$ elements of $\Upsilon$ and $\Phi$. $\widetilde{\Phi}^{(N)}$ is the approximation of the dual frame that determines the SaFARi$^{(N)}$ construction. 
For the ease of notation, we rewrite Eq.~\ref{Xi_partial} as:
\begin{equation}
    \xi = \Upsilon ( \widetilde{\Phi} \Phi - I )
\end{equation}
For a fixed $\Upsilon$, and $\Phi$, we aim to minimize $\| \xi \|^2_F$ with respect to $\widetilde{\Phi}$:
\begin{equation}
    \textit{Argmin}_{\widetilde{\Phi}} \| \Upsilon ( \widetilde{\Phi} \Phi - I ) \|^2_F
\end{equation}
For the optimal $\widetilde{\Phi}$, the partial derivative is zero.
\begin{equation}
    \frac{\partial}{\partial \widetilde{\Phi}} \| \xi \|^2_F = \frac{\partial}{\partial \widetilde{\Phi}} \| \Upsilon ( \widetilde{\Phi} \Phi - I ) \|^2_F = 2 \Upsilon \Upsilon^T ( \widetilde{\Phi} \Phi - I ) \Phi^T =0
\end{equation}
\begin{equation}
\label{optimal_partial_WaLRUS}
   \widetilde{\Phi}= ( \Phi \Phi^T )^{-1} \Phi^T 
\end{equation}
One should note that the described $\widetilde{\Phi}$ is indeed the pseudo-inverse dual for the truncated frame $\Phi_{[0:N]}$. Therefore, among the possible SaFARi$^{(N)}$ constructs for the same frame, the Dual of Truncation (DoT) has optimal representation error. $\square$

\subsection{ Parallelization using the convolution kernel  }

\begin{apptheorem}[\ref{Theorem_scaling_kernel}] \label{Proof_scaling_kernel}
    For SaFARi using the scaling measure, if A is diagonalizable, computing the scaled representation on a sequence with $L$ samples can be done using a kernel multiplication. 
    \\
    a) For the discretization using General Biliniear Transform (GBT) with parameter $\alpha$, the kernel can be computed using:
\begin{equation}
    K_L[i,j]= \frac{\prod_{k=j+1}^{L} \left(1-\frac{1-\alpha}{k+1} \lambda_i \right)}{\prod_{k=j}^{L} \left(1+\frac{\alpha}{k+1} \lambda_i \right)} \in \mathbb{R}  ^{N \times L}
\end{equation}
    b) For long sequences, the kernel K can be approximated using
\begin{equation}
    K_L(i,j)= \frac{1}{j} \left(\frac{j}{L}\right)^{\lambda_i} \in \mathbb{R}  ^{N \times L}
    %\label{eq:Scaling-closed-form-kernel-appendix}
\end{equation}
For either case a or b, the representation is computed as:
\begin{equation}
    c = M K_L \vec{u}, \quad M= V \rm{diag}( V^{-1} B ) 
\end{equation}
where $V$ and $\lambda_i$ are the eigenvector matrix and eigenvalues of $A$.

\end{apptheorem}

\textbf{Proof}: a) rewriting the GBT update rule for the diagnoalzied SSM we have:
\begin{align}
    \widetilde{C}[n+1]&= \left(I + \frac{\alpha}{n+1} \Lambda \right)^{-1} \left(I - \frac{1-\alpha}{n+1} \Lambda \right) \widetilde{C}[n]+  \left(I + \frac{\alpha}{n+1} \Lambda \right)^{-1} \widetilde{B} u[n]
    \\ \nonumber
    &= \bar{A}_n \widetilde{C}[n]+  \bar{B}_n u[n]
\end{align}
\begin{equation}
    \rightarrow \widetilde{C}[L+1]= \bar{B}_L u[L]+ \bar{A}_L \bar{B}_{L-1} u[L-1]+ ...+ \bar{A}_L \dots \bar{A}_1 \bar{B}_{0} u[0]  
\end{equation}
for the ease of computation and notation, we define $K_L$ such that:
\begin{equation}
    K_L[i,j]= \bar{A}_L \dots \bar{A}_{j+1}   \left(1 + \frac{\alpha}{j+1} \lambda_i \right)^{-1}  =\frac{\prod_{k=j+1}^{L} \left(1-\frac{1-\alpha}{k+1} \lambda_i \right)}{\prod_{k=j}^{L} \left(1+\frac{\alpha}{k+1} \lambda_i \right)}
\end{equation}
\begin{equation}
    \widetilde{C}_i= \widetilde{B}_i \sum_j K_L[i,j] u[j]  = \widetilde{B}_i  [K_L \vec{u} ]_i  
\end{equation}
\begin{equation}
   \widetilde{c}= \widetilde{B} \odot (K_L \vec{u}) = \rm{diag}(\widetilde{B}) K_L \vec{u}
\end{equation}
\begin{equation}
   c= V \widetilde{c}= V  \rm{diag}(\widetilde{B}) K_L \vec{u}= M K_L \vec{u} 
\end{equation}

\textbf{Proof}: b) Using Lemma~\ref{Lemma1} for the diagonalized version of Scaled-SaFARi the closed form solution is
\begin{equation}
    \widetilde{c} =  \int_{t=0}^T  \frac{1}{t} \exp{\left( \Lambda \ln \frac{t}{T} \right)}  \widetilde{B} u(t) dt .
\end{equation}
\begin{equation}
    \widetilde{c}_i =  \int_{t=0}^T  \frac{1}{t} \exp{\left( \lambda_i \ln \frac{t}{T} \right)}  \widetilde{B}_i u(t) dt = \widetilde{B}_i \int_{t=0}^T  \frac{1}{t} \left(\frac{t}{T}\right)^{\lambda_i}   u(t) dt   =  \widetilde{B}_i [K_L \vec{u}]_i 
\end{equation}
\begin{equation}
   \widetilde{c}= \widetilde{B} \odot (K_L \vec{u}) = \rm{diag}(\widetilde{B}) K_L \vec{u}
\end{equation}
\begin{equation}
   c= V \widetilde{c}= V  \rm{diag}(\widetilde{B}) K_L \vec{u}= M K_L \vec{u} 
\end{equation}

\begin{apptheorem} [\ref{Theorem_translating_kernel}]  \label{Proof_translating_kernel}
    For SaFARi using translated measure with $\theta_L$ samples long sliding window,if A is diagonalizable, computing the translated representation on a sequence with $L$ samples can be done using a kernel multiplication.
    
    a) for the discretization using General Bilinear Transform(GBT) with parameter $\alpha$, the kernel can be computed using:
    \begin{equation}
    K_L[i,j]=  \frac{1} {1+\frac{\alpha}{\theta} \lambda_i }  \left( \frac{1-\frac{1-\alpha}{\theta} \lambda_i } {1+\frac{\alpha}{\theta} \lambda_i } \right)^{L-j}
    \end{equation}
    
    b) For long sequences, the kernel $K$ can be approximated using
\begin{equation}
    K_L(i,j)= \frac{1}{\theta_L} \exp \left( - \lambda_i \frac{L-j}{\theta_L}  \right)\in \mathbb{R}  ^{N \times L}
    %\label{Translated-closed-form kernel-appendix}
\end{equation}
For either of case a or b, the representation is computed by
\begin{equation}
    c = M K_L \vec{u}, \quad M= V \rm{diag}( V^{-1} B ) \, .
\end{equation}
where $V$ and $\lambda_i$ are the eigenvectors matrix, and eigenvalues of $A$.
\end{apptheorem}

\textbf{Proof}: a) rewriting the GBT update rule for the diagonalized SSM we have:
\begin{equation}
    \widetilde{C}[n+1]= \left(I + \frac{\alpha}{\theta} \Lambda \right)^{-1} \left(I - \frac{1-\alpha}{\theta} \Lambda \right) \widetilde{C}[n]+  \left(I + \frac{\alpha}{\theta} \Lambda \right)^{-1} \widetilde{B} u[n]
    \\
    = \bar{A} \widetilde{C}[n]+  \bar{B} u[n]
\end{equation}
we take a similar approach to the previous theorem. The only difference is that $\bar{A}$ and $\bar{B}$ remain the same for all the time indices. 
\begin{equation}
 \widetilde{C}[L+1]= \bar{B} u[L]+ \bar{A} \bar{B} u[L-1]+ ...+ \bar{A}^L \bar{B} u[0]  
\end{equation}

for the ease of computation and notation, we define $K_L$ such that:
\begin{equation}
    K_L[i,j]=  \frac{1} {1+\frac{\alpha}{\theta} \lambda_i }  \left( \frac{1-\frac{1-\alpha}{\theta} \lambda_i } {1+\frac{\alpha}{\theta} \lambda_i } \right)^{L-j}
\end{equation}

\begin{equation}
    \widetilde{C}_i= \widetilde{B}_i \sum_j K_L[i,j] u[j]  = \widetilde{B}_i  [K_L \vec{u} ]_i  
\end{equation}
\begin{equation}
   \widetilde{c}= \widetilde{B} \odot (K_L \vec{u}) = \rm{diag}(\widetilde{B}) K_L \vec{u}
\end{equation}
\begin{equation}
   c= V \widetilde{c}= V  \rm{diag}(\widetilde{B}) K_L \vec{u}= M K_L \vec{u} 
\end{equation}

\textbf{Proof}: b) Using Lemma~\ref{Lemma2} for the diagonalized version of Translated-SaFARi, the closed-form solution is:
\begin{equation}
    \widetilde{c} = \int_{t=T-\theta}^T \frac{1}{\theta} \exp \left(\Lambda \frac{t-T}{\theta}\right) \widetilde{B} u(t) .
\end{equation}
\begin{equation}
    \widetilde{c}_i = \int_{t=T-\theta}^T \frac{1}{\theta} \exp\left(\lambda_i \frac{t-T}{\theta}\right) \widetilde{B}_i u(t)  =  \widetilde{B}_i [K_L \vec{u}]_i 
\end{equation}
\begin{equation}
   \widetilde{c}= \widetilde{B} \odot (K_L \vec{u}) = \rm{diag}(\widetilde{B}) K_L \vec{u}
\end{equation}
\begin{equation}
   c= V \widetilde{c}= V  \rm{diag}(\widetilde{B}) K_L \vec{u}= M K_L \vec{u} 
\end{equation}

\subsubsection{Numerical instability of the fast sequential legS solver}
\label{section:fast_legs_failure}
As part of our experimental findings, we realized that the proposed method for sequential updates for LegS SSM( in \citet{gu2020hippo},Appendix E) suffers from numerical instability when working with larger SSMs.
\begin{equation}
    x= \frac{\text{cumsum}(\beta \text{cumprod} \frac{1}{\alpha})}{  \text{cumprod} \frac{1}{\alpha} }
\end{equation}

where the introduced $\alpha_k= \frac{d_k}{1+d_k}$ is a decreasing function. One can confirm that in the $t^{\text{th}}$ iteration, and for the $k^{\text{th}}$ degree Legendre polynomial
\begin{equation}
    \alpha_k= \frac{d_k}{1+d_k} = \frac{2(t+1)-k}{  2(t+1) + k + 1 }
\end{equation}
Then the proposed solution requires finding cumulative product of $\frac{1}{\alpha_k}$ for $k \in [1,N]$ in each step.
\begin{equation}
    \log \left|\text{cumprod}_{k'} \left(\frac{1}{\alpha_{k'}}\right)(k)\right|= \sum_{k'=1}^K \log \left| \frac{1}{\alpha_{k'}} \right| = \sum_{k'=1}^K \log\left| -1+ \frac{4(t+1)+1}{2(t+1)-k'}\right|  
\end{equation}
\begin{equation}
     = \sum_{k'=1}^{2(t+1)} \log\left(   -1+ \frac{4(t+1)+1}{2(t+1)-k'}\right)  +\sum_{k'=2(t+1) +1 }^K \log\left(   1+ \frac{4(t+1)+1}{k'-2(t+1)}  \right)  
\end{equation}
\begin{equation}
     = \sum_{k'=1}^{2(t+1)} \log \left( -1+ \frac{4(t+1)+1}{2(t+1)-k'}\right)  +\sum_{k'=1 }^{K-2(t+1)} \log \left(   1+ \frac{4(t+1)+1}{k'}  \right)  
\end{equation}

\begin{figure}
    \includegraphics[width=.5\textwidth]{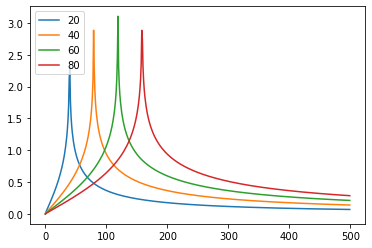}\hfill
    \includegraphics[width=.5\textwidth]{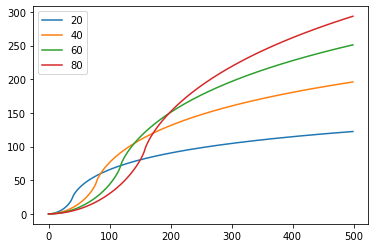}
    \caption{Fast Legs numerically diverges. \textbf{left:} For a system with $N=500$, $\log|\frac{1}{\alpha_k}|$ for different values of $k \in [ 1,N] $ is plotted at different iterations $t=20,40,60,80$. 
   \textbf{right:} log of the cumulative product which is equal to the cumulative sum of the left plot is plotted for different iterations. In the right plot, it is notable that for $t=80$, the cumulative product reaches to $10^{300}$ for a $k<500$ which is the largest value that a float-64 variable can handle. The studied  Fast-LegS method for an SSM having more then 500 coefficients diverges after only $80$ sequential updates.}\label{fig:Fast_legs_failure}
\end{figure}

 For any specific iteration (fixed $t$), as K grows(higher representation index), the second summation above grows to infinity. 
 Thus, for large enough N, $\text{cumprod}(\frac{1}{\alpha_k})$ diverges beyond machine precision. As a result, the proposed fast sequential legS solver proposed in (\citet{gu2020hippo}, Appendix E) fails. Figure.~\ref{fig:Fast_legs_failure} Shows an example where for $N=500$, fast LegS numerically diverges for any sequence longer than $80$ samples. It is crucial to note that this numerical instability is fundamental to legS, and does not depend on the input signal at all.
 
\begin{figure}
    \centering
    \includegraphics[width=.6\textwidth]{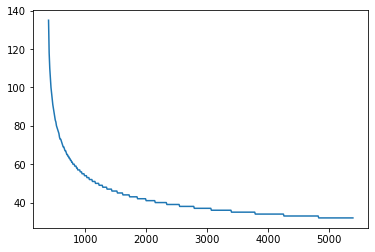}
    \caption{As the given LegS size $N$ grows, the longest sequence length before observing numerical diversion is given in the above plot. For $N<400$ we did not observe the numerical diversion. for $N>400$ , the fast legS sequential update method cannot handle sequences longer than a limited length before it becomes numerically unstable. }\label{fig:Fast_legs_failure2}
\end{figure}

We also investigate the longest sequence length that the given fast legS sequential solution can handle without numerical diversion. Figure~\ref{fig:Fast_legs_failure2} shows that as $N$ grows, then length of sequence that fast legS sequential solver can handle before becoming numerically unstable decrease to a limited length.

The stable version of solving legS  would be to take the similar approach as the fast legS, but in the last step, instead of introducing the proposed $\alpha$ and $\beta$, we find $x_1$, then recursively find $x_i$ after finding all the pervious $x_i$s. This way, the overall computation complexity remains the same, while the run-time complexity becomes longer as one has to compute $x_i$ after $x_0,\dots, x_{i-1}$ are all computed.

\end{document}